\documentclass[lettersize,journal]{IEEEtran}

\usepackage{amsthm}

\usepackage{times}
\usepackage{cite}
\usepackage{multicol}
\usepackage[bookmarks=true]{hyperref}
\usepackage{bm}
\usepackage{amsmath,amsfonts}
\usepackage{algorithm}
\usepackage[noend]{algpseudocode} %
\usepackage{color}
\usepackage{xspace}
\usepackage{graphicx}
\usepackage{makecell}
\usepackage{float}
\usepackage{stfloats}
\usepackage[caption=false,font=footnotesize,labelfont=rm,textfont=rm]{subfig}
\usepackage{multirow}
\usepackage{balance}
\newtheorem{theorem}{Theorem}
\newtheorem{definition}{Definition}
\newtheorem{remark}{Remark}
\newtheorem{assumption}{Assumption}
\newtheorem{problem}{Problem}
\newtheorem{lemma}{Lemma}

\newtheorem{example}{Example}

\usepackage{adjustbox}
\usepackage{comment}
\newcommand{\citep}{\cite}

\newcommand\abbrETO{ETO\xspace}
\newcommand\abbrESDF{ETO-DF\xspace}

\newcommand\abbrMethod{FEO\xspace}
\begin{document}

\title{Ergodic Trajectory Planning with Dynamic Sensor Footprints}

\author{Ziyue~Zheng$^{1,2}$,
        Yongce~Liu$^{1}$,
        Hesheng~Wang$^{3}$,~\IEEEmembership{Senior~Member,~IEEE,}
        and~Zhongqiang~Ren$^{1\dagger}$,~\IEEEmembership{Member,~IEEE}%
\thanks{This work was supported by the Natural Science Foundation of China under Grant 62403313.
\emph{(Corresponding author: Zhongqiang Ren. email: zhongqiang.ren@sjtu.edu.cn)}}%
\thanks{Ziyue Zheng is with Shanghai Jiao Tong University, Shanghai 200240, China, and also with the Department of Mechanical Engineering, Johns Hopkins University, Baltimore, MD 21218 USA.}%
\thanks{Yongce Liu and Zhongqiang Ren are with the Global College, Shanghai Jiao Tong University, Shanghai 200240, China.}%
\thanks{Hesheng Wang is with the School of Automation and Intelligent Sensing, Shanghai Jiao Tong University, Shanghai 200240, China.}}

\markboth{IEEE Transactions on Robotics}%
{Zheng \MakeLowercase{\textit{et al.}}: Ergodic Trajectory Planning with Dynamic Sensor Footprints}

\maketitle

\begin{abstract}
This paper addresses the problem of trajectory planning for information gathering with a dynamic and resolution-varying sensor footprint. Ergodic planning offers a principled framework that balances exploration (visiting all areas) and exploitation (focusing on high-information regions) by planning trajectories such that the time spent in a region is proportional to the amount of information in that region. Existing ergodic planning often oversimplifies the sensing model by assuming a point sensor or a footprint with constant shape and resolution. In practice, the sensor footprint can drastically change over time as the robot moves, such as aerial robots equipped with downward-facing cameras, whose field of view depends on the orientation and altitude. To overcome this limitation, we propose a new metric that accounts for dynamic sensor footprints, and propose efficient numerical trajectory optimization algorithms with computational reuse based on analytical derivation of sensor footprint models. Experimental results show that the proposed approach can simultaneously optimize both the trajectories and sensor footprints, with up to an order of magnitude better ergodicity than conventional methods. We deploy our approach on real robots for verification.

\end{abstract}


\section{Introduction}
\label{sec:intro}
\graphicspath{{figures/}}
This paper investigates a trajectory planning problem for area search, which arises in applications such as search and rescue~\cite{liu2013robotic} and target localization~\cite{mavrommati2017real}.
Given an information map and a probability distribution in the area to be searched, the problem requires planning a trajectory to gather information from this information map efficiently.
Existing approaches to solve this problem span a spectrum that ranges from information-theoretic approaches~\cite{julian2012distributed,chen2021pareto}, which greedily move the robot to the next location with the highest information gain, to uniform coverage methods~\cite{santos2018coverage,julian2012distributed}, which guarantee complete exploration of the entire information map. 
In the middle of this spectrum lies the ergodic search~\cite{mathew2011metrics,miller2013trajectory,de2016ergodic}, which plans trajectories by minimizing an ergodic metric so that, along the planned trajectory, the amount of time spent in a region is proportional to the amount of information in that region.
Ergodic search provides a framework to inherently balance between exploration (i.e., covering all possible locations for new information) and exploitation (i.e., myopically searching high-information areas) and is thus able to intelligently plan the robot motion to gather information in the long run.

\begin{figure}[tb]
\centering
    \includegraphics[width=\linewidth,trim=1 1 1 1,clip]{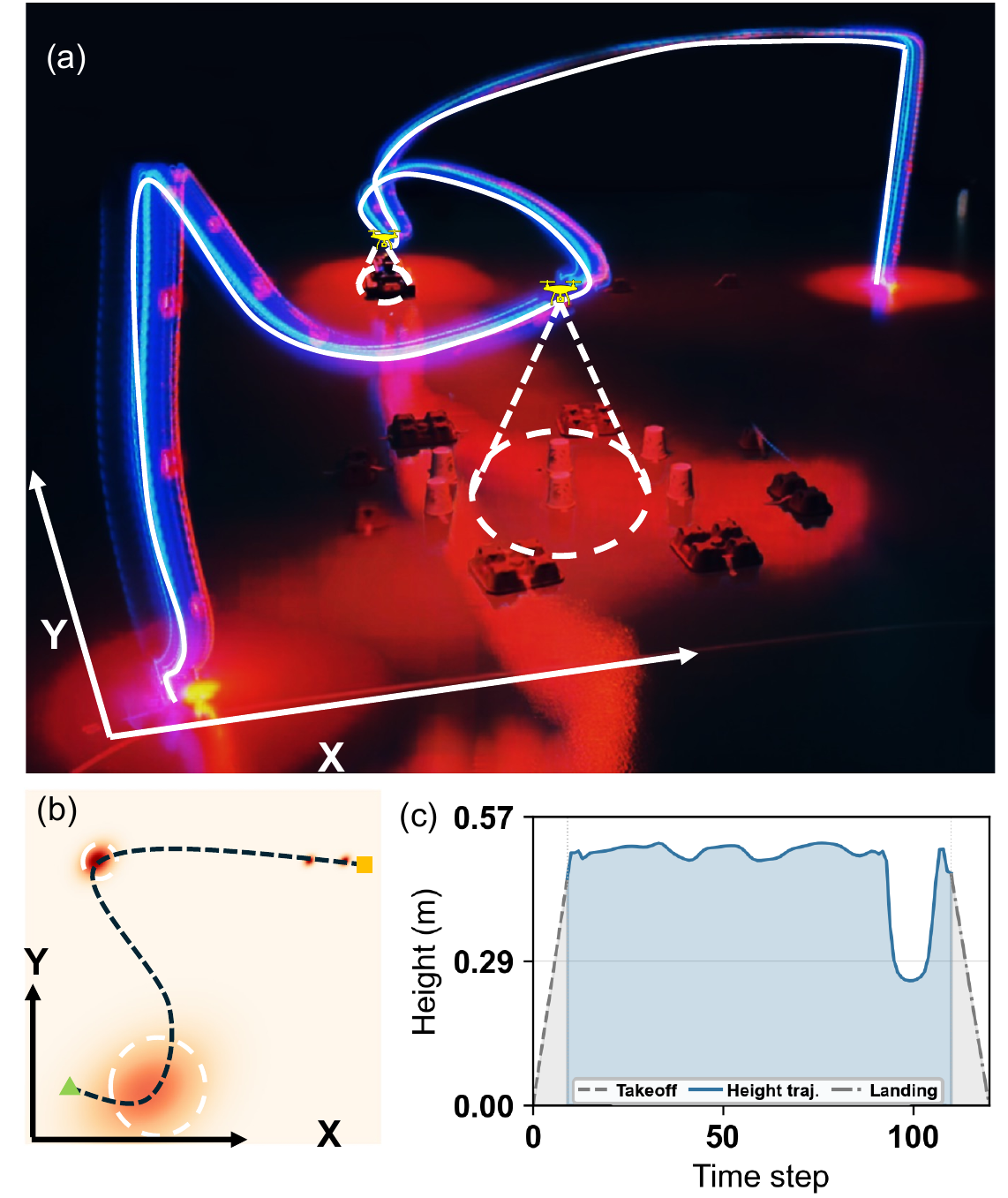}
    \caption{Ergodic trajectory planning with dynamic sensor footprint. (a) shows our real-robot experiments, where a drone covers an information map under a motion capture system. The drone is equipped with a downward pointing LED to visualize the dynamic sensor footprint whose size depends on the flying height of the robot. With our approach, the robot flies high over regions with widespread information and flies low over regions with concentrated information.
    (b) shows the information map to be covered, and the top-down view of the trajectory. (c) shows the corresponding height values (z coordinate) of the robot along the trajectory. The orange shadow on the ground in (a) shows the time-averaged statistics of the dynamic sensor footprint, which is similar to the information map to be covered as shown in (b).}
    \label{fig:abs}
\end{figure}

Early research on ergodic search simplifies the sensor detection range (i.e., sensor footprint), which describes the footprint as either a single point~\cite{mathew2011metrics,miller2013trajectory,abraham2018decentralized} in the workspace or as constant size geometry such as circles or rectangles~\cite{ayvali2017ergodic}.
In practice, the sensor footprint often varies its shape and size as the robot moves and such variation is coupled with both the robot states such as position and orientation, and the space to be covered such as non-flat surface or 3D objects.
For example, consider a drone with a camera whose optical axis points downward and perpendicular to the ground to collect information (Fig.~\ref{fig:abs}).
The field of view (FOV) forms a cone.
The sensor footprint projected onto the ground is a circle whose radius changes with the drone’s altitude.
Moreover, the footprint size correlates negatively with the sensor resolution:
At higher altitudes, the footprint is larger and the resolution is coarser, which collects rough information over a broad area;
At lower altitudes, the footprint is smaller and the resolution is finer, which gathers detailed information within small regions.
When the ground is a non-flat surface, sensor footprint geometry can be more complex, which plagues the trajectory planning.
Ergodic search with varying sensor footprint (or end-effector footprints) on non-flat surfaces is a rising topic recently, for the purpose of surface finishing~\cite{11288096}, full-body exploration~\cite{bilaloglu2023whole,bilaloglu2024tactile} and 3D object inspection~\cite{IVIC2023104709,kwon2026volumetricergodiccontrol}, to name a few.
However, these approaches usually simplify the dependency of footprint geometry, and the corresponding derivatives, on both robot dynamics and the surface to be covered, which may then affect and burden the underlying trajectory optimization.

To capture these dynamic sensing characteristics and their influence on the trajectory optimization, this paper seeks to integrate a dynamically changing sensor footprint and resolution into the ergodic search framework.
In particular, we first focus on 2D non-flat surfaces with bounded curvature (such as hills) covered by a view cone of a robot flying in SE(3).
These surfaces can often be described by manifolds that are topologically equivalent to a 2D plane.
The major challenge there is deriving the geometry of the sensor footprint when the robot flies in SE(3), which leads to oblique views, and its Jacobian with respect to the robot pose, which is often necessary for efficient trajectory optimization.

To address the challenges, first, we introduce a new footprint ergodic metric which computes the time-averaged statistics of the dynamic sensor footprints and compares it against the information map to be covered.
The proposed footprint ergodic metric converges to the existing ergodic metric~\cite{mathew2011metrics} as the sensor footprint approaches a single point in the limit.
Then, with the proposed footprint ergodic metric, we formulate a corresponding ergodic optimal control problem (OCP).
Directly solving this OCP is difficult, since the footprint is a set of infinitely many points and the proposed footprint ergodic metric involves an integration over the set which can be inefficient to compute.
We thus propose a numerical method to compute it approximately based on a finite number of samples within the set.
As the number of samples increases and approaches infinity, our numerical approximation converges to the true footprint ergodic metric.
To incorporate more sophisticated constraints from practice, we leverage both the numerical footprint ergodic metric and the Augmented Lagrangian within an iLQR framework to plan the trajectories, and we name this approach Footprint Ergodic Optimization (\abbrMethod).

Our \abbrMethod involves two novel ideas. 
First, we approximate smooth surface to be covered with local tangent planes at the intersection points of the sensor optical axis with the surface as the robot moves, and derive the analytical expression of the Jacobian of the position of any point within the sensor footprint on this tangent plane with respect to the robot pose in SE(3).
These analytical expressions give us the insight that the sensor footprint with respect to the robot position is inherently \emph{linear}.
Intuitively, it aligns with the fact that, the sensor footprint grows proportionally larger or smaller as the robot translates in 3D (e.g. flying lower or higher even with oblique views).
Second, this linearity suggests that, parts of these Jacobians remain constant during trajectory optimization, which enables \emph{computational reuse} across iterations and helps reduce the runtime of each iteration during the trajectory optimization.
Following this idea, we further look into the computation and leverage computational reuse to speed up our \abbrMethod.
Besides, in \abbrMethod, we also propose a log-spaced line search approach to replace the popular Armijo line search~\cite{Armijo1966MinimizationOF} that is widely used in trajectory optimization, which helps reduce the number of iterations of optimization.

For verification, we compare our \abbrMethod on various flat and non-flat surfaces in simulation against several baselines, including conventional ergodic search~\cite{mathew2011metrics}, ergodic search with fixed Gaussian footprint~\cite{ayvali2017ergodic}, extension of fixed Gaussian footprint to meshable surfaces~\cite{dong2025ergodic}, and the recent HEDAC for complex structures~\cite{IVIC2023104709} and non-planar surfaces \cite{bilaloglu2024tactile,bilaloglu2023whole}.
Our approach often achieves up to more than an order of magnitude smaller (better) ergodicity due to the consideration of dynamic sensor footprint, while running fast due to the proposed computational reuse and log-spaced line search, which speeds up our \abbrMethod by up to 70\%.
Intuitively, our \abbrMethod allows the drone to intelligently fly further away from the surface in regions with low-density information, and closer to the surface in areas with concentrated information. 
We deploy our approach with a drone and motion capture in a lab setting for ergodic search with dynamic sensor footprint.

\section{Related Work}
\label{sec:related}
\subsection{Ergodic Search}
Ergodic search plans trajectories by minimizing the difference between the robot's time-averaged statistics and the workspace's information distribution.
Several approaches have been proposed to evaluate the difference, including spectral multi-scale coverage using Fourier analysis~\cite{mathew2011metrics}, Kullback–Leibler divergence with Gaussian mixture models~\cite{abraham2021ergodic}, kernel ergodic metric for Lie group spaces~\cite{sun2024fast}, and maximum mean discrepancy metric~\cite{hughes2024ergodic}.
For trajectory optimization, gradient-based methods are widely adopted. 
Some techniques involve feedback control for dynamic systems~\cite{mathew2011metrics}, receding-horizon planning~\cite{mavrommati2017real}, and LQR-based optimization~\cite{miller2013trajectory}.
Recent developments have incorporated additional considerations like time efficiency~\cite{dong2024time}, multi-objective criteria~\cite{ren2023pareto}, and inter-robot connectivity~\cite{2025_IROS_ESPC_YongceLiu}.

\subsection{Ergodic Coverage of 3D Targets}
Although being generic to the general workspace, most experiments on ergodic search are limited to a 2D workspace~\cite{mathew2011metrics,abraham2018decentralized}.
Recently, some work investigates coverage in 3D workspaces,
including Multi-UAV trajectory planning for complex structures~\cite{IVIC2023104709}
and coverage of non-planar surfaces in object inspection or surface scanning with robotic arms~\cite{dong2025ergodic,bilaloglu2024tactile}.
Some of these methods extend the ergodic metric~\cite{mathew2011metrics} to handle manifolds, which enables coverage of non-planar surfaces of a 3D target object~\cite{dong2025ergodic}.
Another approach HEDAC adopts a diffusion-based strategy, where ergodic trajectories are generated by solving a heat equation defined over the domain of interest~\cite{IVIC2023104709, bilaloglu2024tactile,bilaloglu2023whole}.
Although general and versatile, this approach requires solving PDE repetitively, which can be computationally expensive~\cite{dong2025ergodic}.

\subsection{Dynamic Sensor Footprints}\label{DSF}
Information gathering problems were investigated in many ways, such as sensor coverage for planning \cite{davis2016c,mostegel2021overlap}, sensor view points in SLAM, target detection and tracking \cite{best2019dec,patten2015viewpoint}.
However, most of them simplify the sensor as a point in the workspace and ignore the detection range and resolution of the sensor~\cite{tang2005motion}, or as a fixed geometry such as
rectangles~\cite{coffin2022multi}, or spheres~\cite{wang2024sensor} without considering that the sensor detection range and resolution vary with robot dynamics.
Other works consider varying sensor footprints and focus mainly on uniform coverage~\cite{sadat2015fractal,galceran2012efficient} or environmental monitoring with different resolutions~\cite{stache2023adaptive}.
Recent research considers varying footprints or non-flat surfaces in tactile coverage~\cite{bilaloglu2024tactile}, surface finishing~\cite{11288096}, 3D object inspection~\cite{IVIC2023104709}
This paper complements these research by investigating the geometry of sensor footprint in 3D, the resulting gradient and Jacobian, and their use for fast computation for ergodic trajectory optimization.

\section{Preliminaries}
\label{sec:preli}
\graphicspath{{figures /}}

\subsection{Workspace and Trajectories} \label{Workspace and Traj}

Let $\mathcal{W}=[0,L_1]\times \cdots \times[0,L_\nu],\nu=2,3$ denote a bounded $\nu$-dimensional workspace, which is to be explored by the robot.\footnote{The workspace in this paper is simply the space of information to be explored by the robot, as opposed to the space where the robot can move. In other words, the robot can move in a higher-dimensional space than the workspace. For example, the robot can be a drone flying in 3D space while exploring the information in a 2D workspace. The primary focus of this work are 2D workspace, i.e., 2D information maps. Later in Sec. \ref{sec:terrain_3d} about coverage of a general 2D manifold embedded in 3D, the underlying information map is still 2D. We discuss how the proposed techniques can be potentially extended to cover a truly 3D information map in Sec.~\ref{sec:conclusion}.}
The mobile robot (such as a drone) has an $n$-dimensional state space ($n\geq \nu$), and let $x:[0, T]\rightarrow\mathbb{R}^n$ denote a trajectory in the state space with $T\in\mathbb{R}^+$ representing the time horizon.
At any time $t \in [0, T]$, the state of the robot $x(t)$ must stay within the set $\mathcal{X} \subseteq \mathbb{R}^{n}$, which is a set of allowed states.
Here, $\mathcal{X}$ describes the constraints on the robot's state, such as avoiding collision with the static obstacles in the workspace or staying within the bounded workspace.
The robot has deterministic dynamics given by $\dot{x}(t)=f_d(x(t),u(t))$, where $u(t) \in \mathcal{U}$ is the control input of the robot.
Let $f_q: \mathbb{R}^n \rightarrow \mathcal{W}$ denote a map that projects a state $x(t)$ to the corresponding point $q(t)\in \mathcal{W}$ in the workspace, i.e., $q(t) = f_q(x(t))$.

\subsection{Ergodic Metric}

Let $\phi(w): \mathcal{W} \rightarrow \mathbb{R}_0^+$ denote an information map over the workspace, which is a time-invariant probability distribution function with $\int_{\mathcal{W}_{-}} \phi(w) dw = 1$, describing the information density at each point in the workspace $\mathcal{W}$. 
Based on the trajectory $q(t)$, the time-averaged statistics for the robot is defined as follows \cite{mathew2011metrics}.
\begin{align}
    \label{eq:time_averaged_statistic}
    c(w,x(t)) = c(w, f_q(x(t))) = \frac{1}{T} \int_{0}^{T} \delta(w-q(t)) dt
\end{align}
where $\delta(w)$ is the Dirac delta function such that $\delta(\mathbf{0})=+\infty$ and $\delta(w)=0, w \ne \mathbf{0}$, satisfying $\int_{\mathcal{W}_{-}} \delta(w) dw =1$.

With the information distribution $\phi(w)$ and the time-averaged statistics (\ref{eq:time_averaged_statistic}), the ergodic metric (ergodicity) is defined as follows~\cite{mathew2011metrics}.
\begin{align}
    \label{eq:ergodic_metric}
    &\mathcal{E}(\phi, q(t)) = \sum_{\mathbf{k} \in \mathcal{K}}\Lambda_\mathbf{k}{\left(c_\mathbf{k}-\phi_\mathbf{k}\right)}^2\\
    &=\sum_{\mathbf{k} \in \mathcal{K}} \Lambda_\mathbf{k} {\left( \frac{1}{T}\ \hspace{-2pt} \int_0^T F_\mathbf{k}(q(t))dt \hspace{-2pt} - \hspace{-3pt} \int_{\mathcal{W}_{-}} \phi(w)F_\mathbf{k}(w)dw \right) }^2.\notag
\end{align}
Here, $c_\mathbf{k}$ and $\phi_\mathbf{k}$ are the Fourier coefficients of $c(w)$ and $\phi(w)$, respectively.
$\mathbf{k}=(k_1, \cdots, k_\nu) \in \mathcal{K}$ is the frequency vector of the Fourier coefficients, and $\mathcal{K} \subset \mathbb{N}^\nu$ represents a selected set of frequencies in practical computation.
Additionally, $F_\mathbf{k}(w)=\frac{1}{h_\mathbf{k}}\prod_{o=1}^{\nu}\cos{\frac{k_o \pi}{L_o}w_o}$ is the cosine basis function with the normalization term $h_\mathbf{k}$~\cite{mathew2011metrics,dong2024time,miller2015ergodic}.
$\Lambda_\mathbf{k}=(1+{\parallel \mathbf{k} \parallel}^2_2)^{-{(\nu+1)}/{2}}$ is the weight of each Fourier coefficient.

\subsection{Ergodic Trajectory Optimization Problem}

Given an information map $\phi$, the goal of the Ergodic Trajectory Optimization (\abbrETO) problem is to find a dynamically feasible trajectory $x$ of the robot that minimizes the ergodic metric $\mathcal{E}(\phi,q)$ and control effort:
\begin{subequations}
\label{prob:es}
\begin{align}
    \min_{x,u} \;\;& \mathcal{E}(\phi,q) + \int_0^Tu^T(t) R u(t)dt \label{prob:es:objective}\\
    \text{s.t. } \dot{x} &= f_d(x(t),u)\label{prob:es:dynamics} \\
    x &\in \mathcal{X} ,u \in \mathcal{U} \label{prob:es:x_u_space}
\end{align}
\end{subequations}
where $R$ is a positive definite matrix describing the weight of the controls.
Constraint \ref{prob:es:dynamics} represents the dynamics of the robot, and \ref{prob:es:x_u_space} defines the constraints of state and control.

\section{Footprint Ergodic Metrics}
\label{sec:method}
\graphicspath{{figures/}}

\subsection{Sensor Footprint and Trajectory}\label{sec:fem}

\begin{definition}[Sensor Footprint]\label{def:sensor_footprint}
Let $\gamma(w,x) \subset \mathcal{W}, w\in \mathcal{W}$ denote the sensor footprint of the robot (over the information map in the workspace) when the robot state is $x$, satisfying $\int_{\mathcal{W}}\gamma(w,x)dw = 1$ and $\gamma(w,x) > 0, \forall w \in \mathcal{W}$.
\end{definition}
Intuitively, when the robot state is $x$, for any point in the workspace $w\in \mathcal{W}$, $\gamma(w,x)$ returns the ``weight of measurement'' at point $w$, and the weight integrated at all possible points $w\in \mathcal{W}$ should be equal to one.
In other words, $\gamma(w,x)$ can be regarded as a probability distribution over the workspace $\mathcal{W}$ that depends on the parameter $x$, and $\gamma(w=w',x(t_k))$ is the value of the probability density function at a specific point $w'\in \mathcal{W}$. 

\begin{figure}[tb]
\centering
    \includegraphics[width=0.7\linewidth, trim=2 1 1 1,clip]{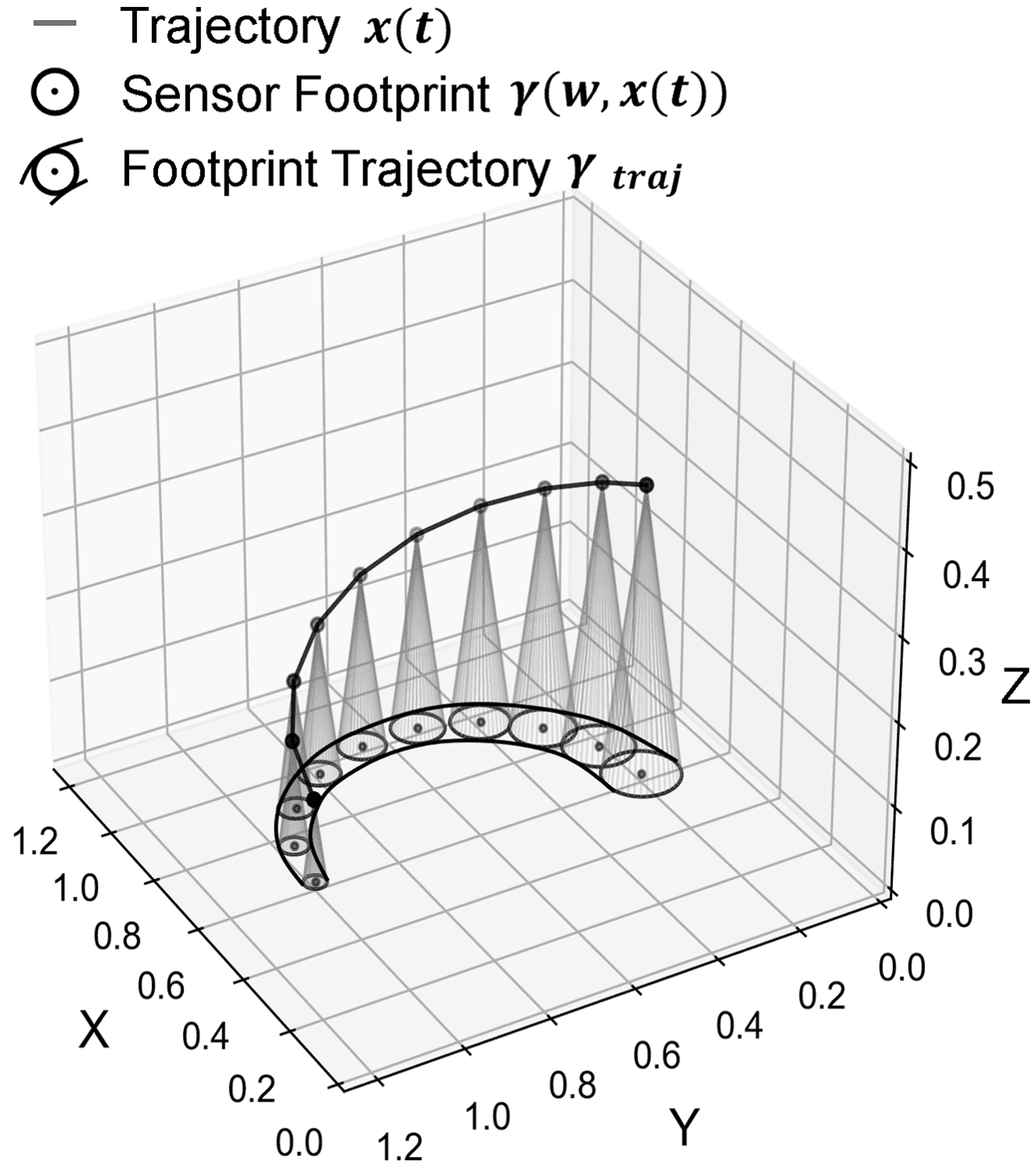}
    \caption{Illustration of sensor footprint and footprint trajectory of a flying drone with downwards pointing camera over a 2D information map.}
    \label{fig:foottraj}
\end{figure}

\begin{example}\label{example1}
Consider a 2D workspace ($\nu=2$) to be explored by a flying robot with a downward pointing camera whose optic axis is always perpendicular to the workspace. 
The sensor footprint can be described by a uniform circle, whose radius depends on the flying height of the robot.
Specifically, for each robot state trajectory $x$, let $f_h:\mathbb{R}^n\rightarrow\mathbb{R}^{+}$ denote a map that projects a state $x(t)$ to a positive real number $h(t)$ representing the height of the robot, i.e., $h(t) = f_h(x(t))$.
\label{sec:ex}
Then, the sensor footprint $\gamma(w,x)$ can be described by the following uniform distribution within the circle (Fig.~\ref{fig:foottraj}).
\begin{align}\label{eq:gam}
    \gamma(w,x(t)) = \begin{cases}
\frac{1}{\pi k_h^2 f_h(x)^2} &\mbox{if } ||w-f_q(x)||_2 \leq k_h f_h(x)  \\
0 & \mbox{otherwise }
\end{cases}
\end{align}
where $k_h \in \mathbb{R}^+$ is a known constant related to the downward viewing angle of the robot.\footnote{To simplify the presentation, this work assumes the entire footprint is always inside the workspace, i.e., $f_q(x)$ is inside $\mathcal{W}$ and is at least $kf_h(x)$ away from the nearest boundary of $\mathcal{W}$.
In practice, this can be ensured by adding constraints on the robot's state as described by $x\in \mathcal{X}$.}
\end{example}

When the robot moves along a trajectory $x(t), t\in [0, T]$, the corresponding sensor footprint varies along $x(t)$.
\begin{definition}[Footprint Trajectory (FT)]\label{def:sft}
Let $\gamma(w,x(t))$ denote a \emph{sensor footprint trajectory}, where for each time point $t_k \in [0,T]$, $\gamma(w,x(t_k))$ is a sensor footprint.
\end{definition}
Fig.~\ref{fig:foottraj} provides an illustration of footprint trajectory over a 2D information map.
Since at any time point $t_k\in[0, T]$, $\gamma(w,x(t_k))$ is a probability distribution over the workspace, an FT $\gamma(w,x(t))$ is actually a \emph{random process}.
We will sample from this random process to develop a numerical approach for easy computation as presented later.

\subsection{Footprint Ergodic Metrics and Problems}

Similar to the time-averaged statistics of a robot trajectory $c(w,x)$ as mentioned in Eq.~\ref{eq:time_averaged_statistic}, we then introduce the time-averaged statistics of an FT $\gamma(w,x(t))$.

\begin{definition}[Time-Averaged Statistics of FT]\label{def:tas_sft}
Let $c_\gamma(w,x(t)), w\in\mathcal{W}, f_q(x(t))\in\mathcal{W}$ denote the time-averaged statistics of an FT $\gamma(w,x(t))$:
\begin{gather}\label{ES_DSF:eqn:time_avg_stat_of_sensor_ftpt_traj}
    c_\gamma(w,x(t)) = \frac{1}{T}\int_{0}^{T} \gamma(w,x(\tau))d\tau.
\end{gather}
\end{definition}
Intuitively, Eq.~\ref{ES_DSF:eqn:time_avg_stat_of_sensor_ftpt_traj} replaces the delta function $\delta$ in Eq.~\ref{eq:time_averaged_statistic} with the footprint trajectory $\gamma$.
To simplify the notation, we omit $t$ in $x(t)$ and write $c_\gamma(w,x(t))$ as $c_\gamma(w,x)$.
An ergodic metric between $c_\gamma(w,x(t))$ and an information map $\phi$ can then be defined in a similar way as in Eq.~\ref{eq:ergodic_metric}.
\begin{definition}[Footprint Ergodic Metric]\label{esdf:eqn:ergodic_metric}
Let $\mathcal{E}_{\gamma}$ denote the \emph{Footprint Ergodic Metric}:
\begin{align}
  &\mathcal{E}_{\gamma}(\phi, x) = \sum_{\mathbf{k} \in \mathcal{K}}\Lambda_\mathbf{k}{\left(c_{\gamma,\mathbf{k}}-\phi_\mathbf{k}\right)}^2 \label{esdf:eq:fem} \\ 
    &=\sum_{\mathbf{k} \in \mathcal{K}} \Lambda_\mathbf{k} {\left(  \int_{\mathcal{W}}
    c_\gamma(w,x) F_\mathbf{k}(w) dw
    \hspace{-2pt} - \hspace{-3pt} \int_{\mathcal{W}} \phi(w) F_\mathbf{k}(w)dw \right) }^2. \notag
\end{align}
Here, $c_{\gamma,\mathbf{k}}$ are the Fourier coefficients of $c_\gamma(w,x)$ for $\mathbf{k} \in \mathcal{K}$.
\end{definition}

The following theorem shows the relationship between the proposed footprint ergodic metric and the conventional ergodic metric~\cite{mathew2011metrics}.
Let $B(\gamma)$ denote a bounding sphere of $\gamma = \gamma(w,x)$ such that for any point $w' \notin B(\gamma)$, $\gamma(w',x)=0$.
In other words, $\gamma$ can only take non-zero values within $B(\gamma)$.
Let $r_B(\gamma)$ denote the radius of $B(\gamma)$.
Intuitively, as $r_B(\gamma)$ approaches zero, since $\int_{\mathcal{W}}\gamma(w,x)dw=1$, the footprint $\gamma(w,x)$ becomes a Dirac delta function at $w=q(x)$.
\begin{theorem}
As $r_B(\gamma)$ approaches $0$:
\begin{align}
    \lim_{r_B(\gamma)\rightarrow0} c_\gamma(w,x(t)) &= c(w,x(t)), \\
    \lim_{r_B(\gamma)\rightarrow0} \mathcal{E}_\gamma(\phi,x) &= \mathcal{E}(\phi,x).
\end{align}
\end{theorem}

\begin{problem}[\abbrESDF Problem]
Given an information map $\phi$, the goal of the Ergodic Trajectory Optimization with Dynamic sensor Footprint (\abbrESDF) problem is to find a dynamically feasible trajectory $x$ of the robot that minimizes the footprint ergodic metric $\mathcal{E}_\gamma(\phi,x)$ and the accumulated control effort:
\begin{align}
\label{eq:obje2}
    \min_{x,u}\, & \mathcal{E}_\gamma(\phi,x) + \int_0^Tu^T(t) Ru(t)dt \\
    \text{s.t. } & (\ref{prob:es:dynamics}),\, (\ref{prob:es:x_u_space}).
\end{align}
\end{problem}

The only difference between \abbrESDF and \abbrETO is to replace the ergodic metric $\mathcal{E}$ in the objective function with the footprint ergodic metric $\mathcal{E}_\gamma$. All other constraints remain unchanged.
The necessary conditions for local optimality for the \abbrESDF problem can be derived in a similar way as in \cite{de2016ergodic,mathew2011metrics, dong2024time, 2025_RSS_IMEC_YongceLiu}, which are omitted here.

\section{Numerical Approach}\label{sec: NA}

\subsection{Approximation of Footprint Ergodic Metric}\label{esdf:sec:num_com_fem}
To solve \abbrESDF numerically, we use direct transcription to convert the continuous-time dynamics and constraints into discrete nonlinear optimization, and then employ an optimization method, such as L-BFGS, to minimize the Lagrange function.
The Lagrange function of \abbrESDF{} can be written as follows, with Lagrange multipliers $\lambda_x(t)\in\mathbb{R}^{n}$:
\begin{align}
\label{eq:alm_func}
\mathcal{L}(x,u,\lambda_x)
= \mathcal{E}_\gamma(\phi,x)
+ \int_{0}^{T}\!\Big[\,
u^{\top}(t)\,R\,u(t) \\ \notag
+\lambda_x^{\top}(t)\big(\dot{x}(t)-f_d(x(t),u(t))\big)
\Big]\,dt,
\end{align}
where $\mathcal{E}_\gamma(\phi,x)$ is the footprint ergodic metric of Def.~\ref{esdf:eqn:ergodic_metric}.

This approach also allows us to handle additional constraints as described in the next section on multi-robot.
To optimize the Lagrangian, the gradient of Eq. \ref{eq:alm_func} with respect to the state $x$ and control $u$ is required:
\begin{align}
\label{meth:LD}
    \frac{\partial \mathcal{L}}{\partial x} = \frac{\partial \mathcal{E}_\gamma(\phi, x)}{\partial x} + \frac{\partial \int_0^T u^TRu + \lambda^T (\dot{x}-f(x, u)) dt}{\partial x}.
\end{align}
The first term $\frac{\partial \mathcal{E}_\gamma(\phi, x)}{\partial x}$ requires an integration over the sensor footprint to compute $\mathcal{E}_{\gamma}$ and then taking the derivative of the integrated value with respect to $x$, which is complicated especially when the sensor footprint $\gamma(w,x)$ corresponds to a complex probability distribution.
We therefore propose an alternative way to numerically compute the footprint ergodic metric $\mathcal{E}_{\gamma}$ approximately.

Let $W_s=\{w_1(t),w_2(t),\cdots,w_M(t)\}$ denote a set of $M$ trajectories in the workspace $\mathcal{W}$ that are realizations of the random process $\gamma(w,x(t))$. Concretely, for any time point $t_k\in [0,T]$, for each realization $w_m\in W_s$, $w_m(t_k)$ is a sampled point from the corresponding random variable $\gamma(w,x(t_k))$ (Fig.~\ref{fig:example}(c)). Here, each sampled trajectory is continuous with respect to time.
Note that, an area with a higher probability value $\gamma(w,x(t_k))$ tends to have more samples $w_m(t_k)$.
Then, a footprint trajectory (FT) in Def.~\ref{def:tas_sft} can be approximated by the following Approximated FT (A-FT):
\begin{align}\label{esdf:eq:approxi_footprint_traj}
    \gamma'(w,x(t)) = \frac{\sum_{m=1}^{M}\delta(w-w_m(x(t)))}{M}.
\end{align}

The time-average statistics of FT in Def. \ref{def:tas_sft} can be approximated with the help of A-FT:
\begin{align}
c'_\gamma(w,x(t)) &= \frac{1}{T}\int_{0}^{T} \gamma'(w,x(\tau))d\tau \nonumber \\
&= \frac{1}{TM}\int_{0}^{T} \sum_{m=1}^{M}\delta(w-w_m(x(\tau))) d\tau
\end{align}
whose corresponding Fourier coefficients are:
\begin{align}\label{esdf:eq:c_prime_gamma}
c'_{\gamma, \mathbf{k}} = \frac{1}{TM} \int_0^T \sum_{m=1}^{M} F_\mathbf{k}(w_m(x(\tau)))d\tau, \mathbf{k} \in \mathcal{K}.
\end{align}
The footprint ergodic metric (Def. \ref{esdf:eqn:ergodic_metric}) can be approximated by:
\begin{align}
  \mathcal{E}'_{\gamma}(\phi, x) = \sum_{\mathbf{k} \in \mathcal{K}}\Lambda_\mathbf{k}{\left(c'_{\gamma,\mathbf{k}}-\phi_\mathbf{k}\right)}^2.
\end{align}

Then, the term $\frac{\partial \mathcal{E}_\gamma(\phi, x)}{\partial x}$ required in Eq.~\ref{meth:LD} can be approximated by $\frac{\partial \mathcal{E}'_\gamma(\phi, x)}{\partial x}$, whose gradient can be derived as follows.

\begin{align}
\nonumber
&\frac{\partial \mathcal{E}'_{\gamma}(\phi, x)}{\partial x}
\nonumber 
=2\sum_{\mathbf{k} \in \mathcal{K}}\Lambda_\mathbf{k}(c'_{\gamma,\mathbf{k}}-\phi_\mathbf{k})\frac{\partial c'_{\gamma,\mathbf{k}}}{\partial x} \\
\nonumber 
&=\frac{2}{TM}\sum_{\mathbf{k} \in \mathcal{K}}\Lambda_\mathbf{k}(c'_{\gamma,\mathbf{k}}-\phi_\mathbf{k})
\int_0^T \sum_{m=1}^{M}\frac{\partial F_\mathbf{k}(w_m(x(\tau)))}{\partial x}d\tau \\
&=\frac{2}{TM}\sum_{\mathbf{k} \in \mathcal{K}}\Lambda_\mathbf{k}(c'_{\gamma,\mathbf{k}}-\phi_\mathbf{k}) \label{eq:a_esdf_gamma}
\\
\nonumber
&\quad\quad \int_0^T \sum_{m=1}^{M}\nabla_w F_{\mathbf{k}}(w_m(x(\tau)))^\top
\frac{\partial w_m(x(\tau))}{\partial x}d\tau
\end{align}
where $\partial w_m(x(\tau))/\partial x$ is the jacobian of the sampled trajectory $w_m(x)$ (from the sensor footprint) with respect to the robot state $x$, and $\nabla_w F_{\mathbf{k}}$ is the gradient of the cosine basis function.
Specifically, we use subscript $j$ to denote the $j$-th component of $\nabla_w F_{\mathbf{k}}(w)$:
\begin{align}
\nonumber
&\bigl[\nabla_w F_{\mathbf{k}}(w)\bigr]_j
=\frac{\partial F_{\mathbf{k}}(w)}{\partial w_j} \\
\nonumber
&=-\frac{1}{h_{\mathbf{k}}}\frac{k_j\pi}{L_j}\sin\!\Bigl(\frac{k_j\pi}{L_j}w_j\Bigr)\prod_{o\neq j}
\cos\!\Bigl(\frac{k_o\pi}{L_o}w_o\Bigr), \; j=1,\dots,\nu.
\end{align}

\subsection{Properties of the Gradients}\label{sec: PotG}
While the gradient $\frac{\partial \mathcal{E}'_\gamma(\phi, x)}{\partial x}$ can be evaluated using Eq.~\ref{eq:a_esdf_gamma} in each iteration of the trajectory optimization, we find a more efficient way to compute Eq.~\ref{eq:a_esdf_gamma} by exploiting the \emph{linearity} in the last term $\frac{\partial w_m(x(\tau))}{\partial x}$ in Eq.~\ref{eq:a_esdf_gamma}, which reduces the runtime complexity.
We re-explained this linearity at the end of this section in Remark \ref{remark:linear}.

\begin{figure}[tb]
    \centering    \includegraphics[width=\linewidth,trim=3 2 2 2,clip]{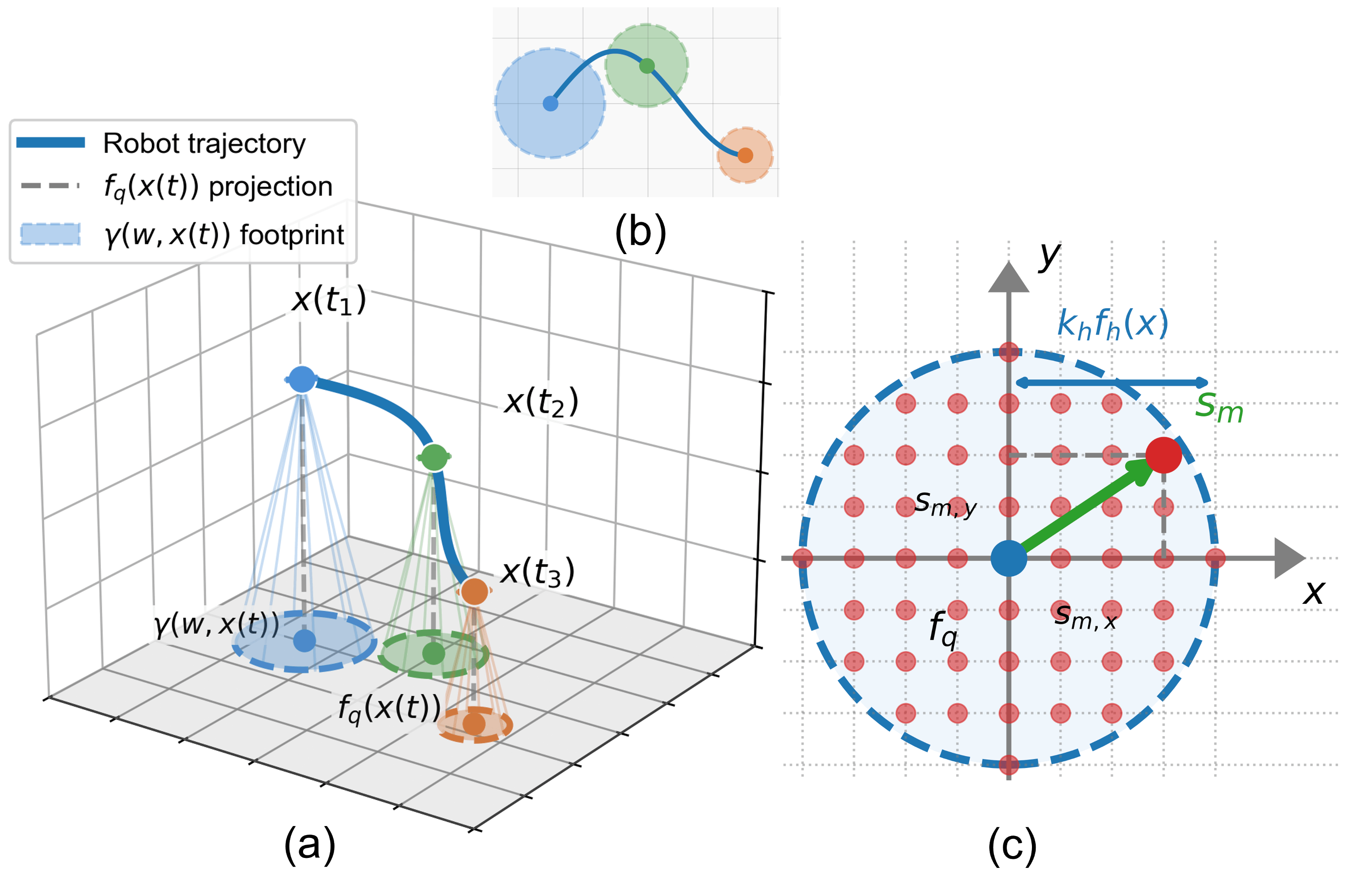}
    \vspace{-5mm}
    \caption{Sampled trajectory $w_m(x(t))$ within the sensor footprint. 
    (a) Robot trajectory $x(t)$ at three time instances with corresponding footprints $\gamma(w,x(t))$ projected onto the workspace. 
    (b) Top-down view of the footprint trajectory. 
    (c) Sampling geometry: directional vector $s_m$ from projected position $f_q(x)$ to sampled point $w_m$, scaled by $k_h f_h(x)$.}
    \label{fig:example}
\end{figure}

Each sampled trajectory $w_m(t)$ in the sensor footprint can be represented by:
\begin{equation}\label{eq:uniform_sample_map}
w_m(x(t)) = f_q(x(t)) + k_h f_h(x(t))\, s_m, m=1,2,\cdots,M
\end{equation}
where $f_q(x(t))$ is the projected position of the robot in the workspace at time $t$ as described in Sec.~\ref{sec:preli} $f_h(x(t))$ is the altitude of the robot at time $t$ as in described in Sec.~\ref{sec:fem}, and $s_m=(s_{m,x},s_{m,y}) \in \mathbb{R}^2$ is the directional (column) vector that points from the robot position to the position of the sampled point when the robot's altitude is one unit (Fig.~\ref{fig:example}). 
Note that, when a deterministic sampling is used, as we do in this work, the vector $s_m$ is a \emph{constant vector} over time, and the distance between the robot position $f_q(x(t))$ and the sampled trajectory point (at time $t$) $w_m(x(t))$ is equal to vector $s_m$ scaled by the altitude of the robot $k_h f_h(x(t))$ (Fig.~\ref{fig:example}).

As a result, the last term $\frac{\partial w_m(x(\tau))}{\partial x}$ in Eq.~\ref{eq:a_esdf_gamma} has the following form
\begin{equation}
\label{eq:wm_general_jac}
\frac{\partial w_{m}(x)}{\partial x}
=
\frac{\partial f_q(x)}{\partial x}
+
k_h\, s_m\,
\Bigl(\frac{\partial f_h(x)}{\partial x}\Bigr).
\end{equation}
Note that the map $f_q(x)$ extracts the 2D position vector of the robot from the robot state $x$.
In practice, the position vector of a mobile robot are usually the first two components of the robot state $x$.
As a result, the first term $
\frac{\partial f_q(x)}{\partial x} \in \mathbb{R}^{2\times n}$ in Eq.~\ref{eq:wm_general_jac} has the following form
\begin{align}
    \frac{\partial f_q(x)}{\partial x} = 
\begin{pmatrix}
1 & 0 & \cdots & 0\\
0 & 1 & \cdots & 0
\end{pmatrix},
\end{align}
which is a constant matrix.

Similarly, the map $f_h(x)$ extracts the altitude, which is for example the third component of the robot state $x$, and the term $\frac{\partial f_h(x)}{\partial x} \in \mathbb{R}^{1\times n}$ in Eq.~\ref{eq:wm_general_jac} has the following form 
\begin{align}
    \frac{\partial f_h(x)}{\partial x} = 
\begin{pmatrix}
0 & 0 & 1 & 0 & \cdots & 0\\
\end{pmatrix},
\end{align}
which is a constant row vector.
As a result, the second term in Eq.~\ref{eq:wm_general_jac} is:
\begin{align}
k_h\, s_m\,
\Bigl(\frac{\partial f_h(x)}{\partial x}\Bigr) = 
\begin{pmatrix}
0 & 0 & k_h s_{m,x} & 0 & \cdots & 0\\
0 & 0 & k_h s_{m,y} & 0 & \cdots & 0
\end{pmatrix}.
\end{align}
Eq.~\ref{eq:wm_general_jac} can be written as:
\begin{align}\label{eq:wm_general_jac_const}
\frac{\partial w_{m}(x)}{\partial x} = 
\begin{pmatrix}
1 & 0 & k_h s_{m,x} & 0 & \cdots & 0\\
0 & 1 & k_h s_{m,y} & 0 & \cdots & 0
\end{pmatrix},
\end{align}
which depends purely on $s_m$.

The set of all $s_m, m=1,2\cdots,M$ can be pre-computed, and as a result, the gradient term $\frac{\partial w_{m}(x)}{\partial x}$ can be pre-computed and reused in all iterations during the trajectory optimization.
\begin{example}[Uniform Sampling in the Euclidean Space]\label{example: USES}
{Let $s_m=(s_{m,x},s_{m,y}) = (k_x\Delta,k_y\Delta)$, where $\Delta$ is a user-specified sampling step sizes along the $x,y$ axes, 
with $k_x,k_y \in \mathbb{Z}, k_x,k_y \in [\frac{-1}{\Delta},\frac{1}{\Delta}]$ }
being the indices of the samples.
Substituting this expression into Eq.~\ref{eq:wm_general_jac_const} yields
\begin{equation}
\frac{\partial w_{m}(x)}{\partial x}=
\begin{pmatrix}
1 & 0 & k_h\, k_x \Delta & 0 & \cdots & 0\\
0 & 1 & k_h\, k_y \Delta & 0 & \cdots & 0
\end{pmatrix}.
\end{equation}
\end{example}
\begin{example}[Uniform Sampling in the Polar Coordinates]\label{example: polar}
Consider a polar grid inside the unit disk with $d_1$ radial levels and $d_2$ angular samples, and let $M=d_1d_2$. For indices $i\in\{1,\dots,d_1\}$ and $j \in\{0,\dots,d_2-1\}$, define
\begin{equation}
r_{i}=\frac{i}{d_1},\qquad
\theta_{j}=\frac{2\pi j}{d_2},
\end{equation}
and set the corresponding samples as
\begin{equation}
s_m=(s_{m,x},s_{m,y})^\top
=
\bigl(r_i\cos\theta_j,\; r_i\sin\theta_j\bigr)^\top.
\end{equation}
Substituting this expression into Eq.~\ref{eq:wm_general_jac_const} yields
\begin{equation}
\frac{\partial w_{m}(x)}{\partial x}=
\begin{pmatrix}
1 & 0 & \, k_h\,r_i\cos\theta_j & 0 & \cdots & 0\\
0 & 1 & \, k_h\,r_i\sin\theta_j & 0 & \cdots & 0
\end{pmatrix}.
\end{equation}
Note that this sampling in polar coordinates corresponds to a non-uniform circular sensor footprint with more weights near the center and with less weights near the boundary, as opposed to the uniform circular sensor footprint as in Example \ref{example1}.
\end{example}

\begin{remark}\label{remark:linear}
    The derivative $\frac{\partial w_m(x(\tau))}{\partial x}$ is constant, which is caused by the fact that the sampled point $w_m(x(t))$ at any time point $t$ is a linear function with respect to $x(t)$ (Eq.~\ref{eq:wm_general_jac}).
    Intuitively, if the robot flies higher (or lower), the sampled point $w_m(x(t))$ moves further away from (or closer to) the center of the sensor footprint proportionally, and if the robot moves along the $\textsc{x,y}$-axes horizontally, the sampled point $w_m(x(t))$ moves along the $\textsc{x,y}$-axes horizontally correspondingly.
\end{remark}

\subsection{Augmented Lagrangian and iLQR}
\label{sec:impl_al_ilqr}

This section presents our \abbrMethod, which is based on Augmented Lagrangian and iLQR numerical optimization approach (Alg.~\ref{alg:al_ilqr}) to solve the \abbrESDF problem.
\abbrMethod leverages that $\frac{\partial w_m}{\partial x}$ is a constant and can be pre-computed and re-used during the iLQR iterations to save computational burden.

\subsubsection{Discrete Problem Formulation}
Let $x_{0:N},u_{0:N-1}$ denote the discretized state and control trajectories and let $x_{k+1} = f_d(x_k,u_k)$ denote the discrete dynamics of the system.
The approximated footprint ergodic metric can be rewritten as:
\begin{equation}\label{ESDF_metriciLQR}
  \mathcal{E}'_{\gamma}(\phi,x_{0:N})
  =\sum_{\mathbf{k}\in\mathcal{K}}\Lambda_{\mathbf{k}}
  \bigl(c'_{\gamma,\mathbf{k}}(x_{0:N})-\phi_{\mathbf{k}}\bigr)^2 .
\end{equation}
The \abbrESDF problem can be rewritten into discrete version as follows.
\begin{problem}[ETO-DF via iLQR Problem]
Based on the approximate footprint ergodic metric $\mathcal{E}'_{\gamma}$, we formulate the following discrete-time optimal control problem:
\begin{align}
\min_{x_{0:N},\,u_{0:N-1}}& \;\; \mathcal{E}'_{\gamma}(\phi,x_{0:N}) +\sum_{k=0}^{N-1} u_k^\top R u_k \\
\text{s.t. } &x_{k+1} = f_d(x_k,u_k), \quad k = 0,\dots,N-1 \\
&x_k \in \mathcal{X} ,u_k \in \mathcal{U}
\end{align}
\end{problem}

\subsubsection{Augmented Lagrangian and Gradients}
To solve this problem, we incorporate these constraints via an augmented Lagrangian $\mathcal{L}_\mathrm{AL}$ of the form
\begin{align}
\nonumber
    \mathcal{L}_\mathrm{AL}(x_{0:N},u_{0:N-1},\boldsymbol{\mu},r)
&=\mathcal{E}'_{\gamma}(\phi,x_{0:N})
+\sum_{k=0}^{N-1} u_k^\top R u_k \\
&+ \Psi(g(x_{0:N},u_{0:N-1});\boldsymbol{\mu},r),
\label{eq:al_merit}
\end{align}
where $g(x_{0:N}, u_{0:N-1})$ represents the aggregate constraint violation, 
$\boldsymbol{\mu}$ denotes the vector of Lagrange multipliers (dual variables), 
$r > 0$ is the penalty parameter, and $\Psi(\cdot)$ is the augmented penalty term 
defined as:
\begin{equation}
\Psi(g; \boldsymbol{\mu}, r) = \frac{1}{2r} \sum_{i=1}^{m} 
\left[ \max(0, \mu_i + r g_i)^2 - \mu_i^2 \right].
\label{eq:penalty_term}
\end{equation}
Unlike indirect methods (Eq.~\ref{eq:alm_func}) where co-state variables enforce equality dynamics constraints in the Lagrangian, we use direct transcription: dynamics are satisfied explicitly through forward integration.
The $\mathcal{L}_\mathrm{AL}$ handles only the inequality constraints via the penalty term $\Psi(\cdot; \boldsymbol{\mu}, r)$. The gradient of $\mathcal{L}_\mathrm{AL}$ with respect to each state and control sample $(x_k,u_k)$ can be written as
\begin{align}
 &\nabla_{x_k}\mathcal{L}_\mathrm{AL}
  =\nabla_{x_k}\mathcal{E}'_{\gamma}(\phi,x_{0:N})+ \nabla_{x_k}\Psi \label{eq:L_AL_x_grad}\\
  &\nabla_{u_k}\mathcal{L}_\mathrm{AL}=2R u_k + \nabla_{u_k}\Psi    
\end{align}
where $\nabla_{x_k}\mathcal{E}'_{\gamma}$ is obtained from the sampling-based expression in Eq.~\ref{eq:a_esdf_gamma}.
\begin{algorithm}[t]
\caption{\abbrMethod}
\label{alg:al_ilqr}
\begin{algorithmic}[1]
\small
\Require Initial guess $x_0$, $\mathbf{u}_{0:N-1}$, penalty $r$, multipliers $\boldsymbol{\mu}$
\Ensure Optimized $\mathbf{x}_{0:N}^*$, $\mathbf{u}_{0:N-1}^*$
\Statex \hspace{-1em} \textit{// Precompute constant Jacobians (reused across all iterations)}
\State Precompute $\{\partial w_m / \partial x\}_{m=1}^M$ via Eq.~\ref{eq:wm_general_jac_const} \label{line:precompute}
\While{not converged}
    \For{$i = 1$ \textbf{to} $N_{\mathrm{inner}}$}
        \State Forward rollout: compute $\mathbf{x}_{0:N}$ from $\mathbf{u}_{0:N-1}$
        \State Linearize dynamics: $A_k, B_k$ at each $(x_k, u_k)$
        \Statex \hspace{2em} \textit{// Gradient computation reuses $\{\partial w_m / \partial x\}$ from Line 1}  \label{line:grad}
        \State Compute $q_k \gets \nabla_{x_k} \mathcal{L}_{\mathrm{AL}}$, $r_k \gets \nabla_{u_k} \mathcal{L}_{\mathrm{AL}}$
        \State Backward pass: compute $K_k, d_k$ via Riccati recursion
        \State Forward pass: $\delta u_k \gets K_k \delta x_k + d_k$, $\forall k$
        \State $\alpha^* \gets$ \Call{LineSearch}{$\mathbf{u}, \boldsymbol{\delta u}, \mathcal{L}_{\mathrm{AL}}$}
        \State $u_k \gets \mathrm{clip}(u_k + \alpha^* \delta u_k, u_{\min}, u_{\max})$, $\forall k$
    \EndFor
    \State Update $\boldsymbol{\mu}$ or increase $r$ based on constraint satisfaction
\EndWhile
\State \Return $\mathbf{x}_{0:N}^*$, $\mathbf{u}_{0:N-1}^*$
\end{algorithmic}
\end{algorithm}
\subsubsection{\abbrMethod Summary}
Our \abbrMethod (Alg.~\ref{alg:al_ilqr}) is similar to the regular iLQR \cite{miller2013trajectory,howell2019altro,2025_RSS_IMEC_YongceLiu}, which employs a double-loop structure.
The inner loop (Lines 3--10) performs iLQR updates: given the current trajectory, we linearize dynamics and compute the augmented Lagrangian gradients $(q_k, r_k)$.
The backward Riccati recursion computes $P_k$ (quadratic cost-to-go) and $r_k$ (linear term), followed by a forward pass that yields the descent direction $\{\delta u_k\}$.
The outer loop (Lines 11--12) updates dual multipliers $\boldsymbol{\mu}$ when sufficient decrease is achieved, or increases penalty $r$ otherwise.

A computational advantage of Alg.~\ref{alg:al_ilqr} is that the terms $\{\frac{\partial w_m}{\partial x}\}$ are precomputed at the beginning (Line 1), and are reused when computing $\nabla_{x_k}\mathcal{E}'_{\gamma}$ in Eq. \ref{eq:L_AL_x_grad} (Line 6), which speeds up the algorithm.
We can do so because $\{\frac{\partial w_m}{\partial x}\}$ depends only on the fixed sample set rather than the evolving trajectory as aforementioned.

\subsubsection{Accelerated Log-Spaced Line Search}
\label{subsec:linesearch} At each \abbrMethod iteration (Section~\ref{sec:impl_al_ilqr}), the backward pass yields a descent direction
$\Delta \mathbf{u}_{0:N-1}:=\{\Delta \mathbf{u}_k\}_{k=0}^{N-1}$ for the current control sequence
$\mathbf{u}_{0:N-1}:=\{\mathbf{u}_k\}_{k=0}^{N-1}$. For a scalar step size $\alpha>0$, we form the candidate update
\begin{equation}
\mathbf{u}_k(\alpha)=\Pi_{\mathcal{U}}\!\left(\mathbf{u}_k+\alpha\,\Delta \mathbf{u}_k\right)
\label{eq:control_update_projection}
\end{equation}
where $\mathcal{U}$ is the admissible control set and $\Pi_\mathcal{U}$ means the projection onto the this control set.
The corresponding state trajectory $\mathbf{x}_{0:N}(\alpha)$ is obtained via a forward rollout.

To select the best step size $\alpha$, a line search \textsc{LineSearch} is needed in each iteration of \abbrMethod, which selects $\alpha$ by minimizing the augmented-Lagrangian merit $\mathcal{L}_\mathrm{AL}$ defined in Eq.~\ref{eq:al_merit}, evaluated on the rolled-out trajectories $\bigl(\mathbf{x}_{0:N}(\alpha),\mathbf{u}_{0:N-1}(\alpha)\bigr)$.
In this \textsc{LineSearch}, multiple candidate values of $\alpha$ are evaluated, each requiring a rollout and recomputation of (i) the sampling-based footprint ergodic objective and (ii) the inequality penalties, making step-size selection expensive if many $\alpha$ are tried.

Classical Armijo backtracking line search accepts the first step size $\alpha$ satisfying the condition 
$\mathcal{L}(\alpha) \leq \mathcal{L}(0) - c \alpha \nabla \mathcal{L}^\top \Delta \mathbf{u}$ during the iterative updates of $\alpha$ \cite{2025_RSS_IMEC_YongceLiu, miller2013trajectory, Armijo1966MinimizationOF}.\footnote{For example, in the $i$-th iteration, $\alpha_i = \alpha_0 \beta^i$, we use $\beta=0.5$ in our implementation.}
If no candidate yields improvement, a predefined smallest step size is used.
Such a process can be inefficient when each merit evaluation requires costly computation.
\label{Armijo}

In Alg.~\ref{alg:al_ilqr}, instead of iterative backtracking, we speed up this \textsc{LineSearch} by evaluating a fixed set $\mathcal{A} = \{\alpha_i\}_{i=0}^{N_\alpha-1}$ of $N_\alpha$ log-spaced step sizes between $\alpha_0$ and $\alpha_{\min}$, where
\begin{align}
\label{eq:log_spaced_steps}
\alpha_i &= \alpha_0 \left(\frac{\alpha_{\min}}{\alpha_0}\right)^{\frac{i}{N_\alpha-1}}.
\end{align}
The best $\alpha^*$ in this set is then selected
\begin{equation}
\alpha^\star = \arg\min_{\alpha \in \mathcal{A}} \;
\mathcal{L}_\mathrm{AL}\!\left(\mathbf{x}_{0:N}(\alpha),\mathbf{u}_{0:N-1}(\alpha),\boldsymbol{\mu},r\right).
\label{eq:alpha_star}
\end{equation}
We accept this $\alpha^*$ if it decreases the current cost $\mathcal{L}_\mathrm{AL}$.
Otherwise, we keep the current controls unchanged. This strategy offers two advantages: (i) the fixed evaluation count enables parallelization, and (ii) selecting the best rather than the first acceptable step often yields larger cost reductions per iteration.
Experimental results show that this log-spaced line search speeds up the computation obviously as detailed later.

\begin{algorithm}[t]
\caption{\textsc{LineSearch}: Accelerated Log-Spaced Search}
\label{alg:log_linesearch}
\begin{algorithmic}[1]
\small
\Require current $\mathbf{u}_{0:N-1}$, direction $\Delta\mathbf{u}_{0:N-1}$, $(\boldsymbol{\mu},r)$, $(\alpha_0,\alpha_{\min},N_\alpha)$
\State Construct candidate set $\mathcal{A}$ using~\ref{eq:log_spaced_steps}
\ForAll{$\alpha \in \mathcal{A}$}
  \State $\mathbf{u}_{0:N-1}(\alpha) \leftarrow \Pi_{\mathcal{U}}\!\left(\mathbf{u}_{0:N-1} + \alpha\,\Delta\mathbf{u}_{0:N-1}\right)$
  \State Roll out $\mathbf{x}_{0:N}(\alpha)$
  \State $J(\alpha) \leftarrow \mathcal{L}_\mathrm{AL}\!\left(\mathbf{x}_{0:N}(\alpha),\mathbf{u}_{0:N-1}(\alpha),\boldsymbol{\mu},r\right)$ using~\ref{eq:al_merit}
\EndFor
\State $\alpha^\star \leftarrow \arg\min_{\alpha\in\mathcal{A}} J(\alpha)$
\If{$J(\alpha^\star) \le J(0)$}
  \State \textbf{return} $\mathbf{u}_{0:N-1}(\alpha^\star)$
\Else
  \State \textbf{return} $\mathbf{u}_{0:N-1}$
\EndIf
\end{algorithmic}
\end{algorithm}

\color{black}

\section{Extensions}
\label{sec:terrain_3d}
While the previous sections assume the robot's FOV is represented by a view cone whose optical axis is always perpendicular to the flat 2D ground, this section allows the robot to fly in SE(3), and the sensor footprint can be projected onto a general surface that is non-flat.
Correspondingly, the view cone may not be perpendicular to the ground and can be oriented in any direction.

\begin{figure}[tb]
    \centering
    \includegraphics[width=\linewidth,trim=8 7 7 7,clip]{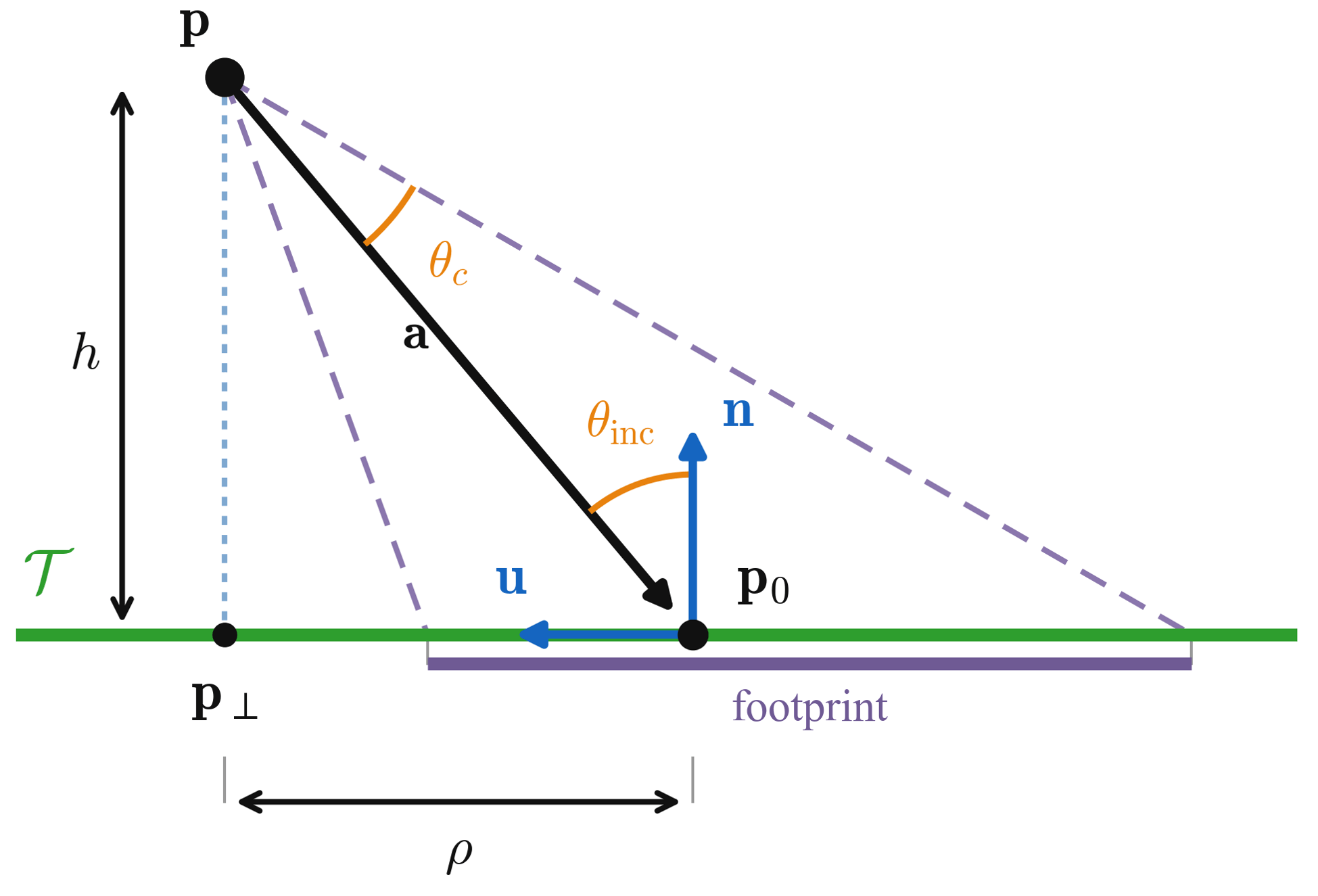}
    \vspace{-5mm}
    \caption{Geometric setup for the 3D footprint model. The view cone with half-angle $\theta_c$ originates from the robot position $\mathbf{p}$ with optical axis $\mathbf{a}$, intersecting the terrain at the contact point $\mathbf{p}_0$. The local tangent frame $\mathcal{L}_T = \{\mathbf{u}, \mathbf{v}, \mathbf{n}\}$ is an orthonormal basis with origin at $\mathbf{p}_0$, where $\mathbf{n}$ is the surface normal, $\mathbf{u}$ points toward the projected robot position $\mathbf{p}_{\perp}$, and $\mathbf{v} = \mathbf{n} \times \mathbf{u}$ points out of the plane of the figure. The incidence angle $\theta_{\mathrm{inc}}$ is the angle between the optical axis and the surface normal.}
    \label{fig:coordinate_schematic}
\end{figure}

For presentation purposes, we begin with the case that the information map remains a flat 2D plane (Fig.~\ref{fig:coordinate_schematic}) while the robot flies in SE(3).
At the end of the section, we present how to generalize the proposed approach to surfaces in 3D by approximating the surface with local tangent planes computed by local Taylor expansions.
When the information map is a 2D flat ground, the local tangent plane at any point is always the same as the information map.
\begin{assumption}\label{assume1}
The derivation in this section assumes that the surface to be covered is smooth, has bounded curvature, and is topologically equivalent to a 2D plane.
\end{assumption}
The purpose of having this assumption is that the surface can then be represented by a single flat information map with little distortion, and that the sensor does not view the surface at near-grazing incidence.
The error introduced by the tangent-plane approximation, as we will introduce soon, is bounded as analyzed in Sec.~\ref{sec:tangent_error}.

We derive both the resulting footprint on the local tangent plane at the point where the optical axis intersects the surface and the Jacobians required for gradient-based trajectory optimization.
In contrast to the 2D case in the previous section, here, only part of the gradient
$\frac{\partial w_{m}(x)}{\partial x}$ remains constant in 3D, while the remaining terms become pose-dependent. 
Computational reuse is still possible for those constant parts within our \abbrMethod framework.

For complex surfaces that violate this assumption, such as closed surfaces that are topologically different from a plane, we briefly discuss how to adapt our approach and showcase its use in Sec.~\ref{3Dreal} in the experimental results, as it is not the focus of our approach.

\subsection{Overview}\label{sec:workspace_mapping}
\begin{figure}
    \centering
    \includegraphics[width=0.9\linewidth,trim=3 3 3 3,clip]{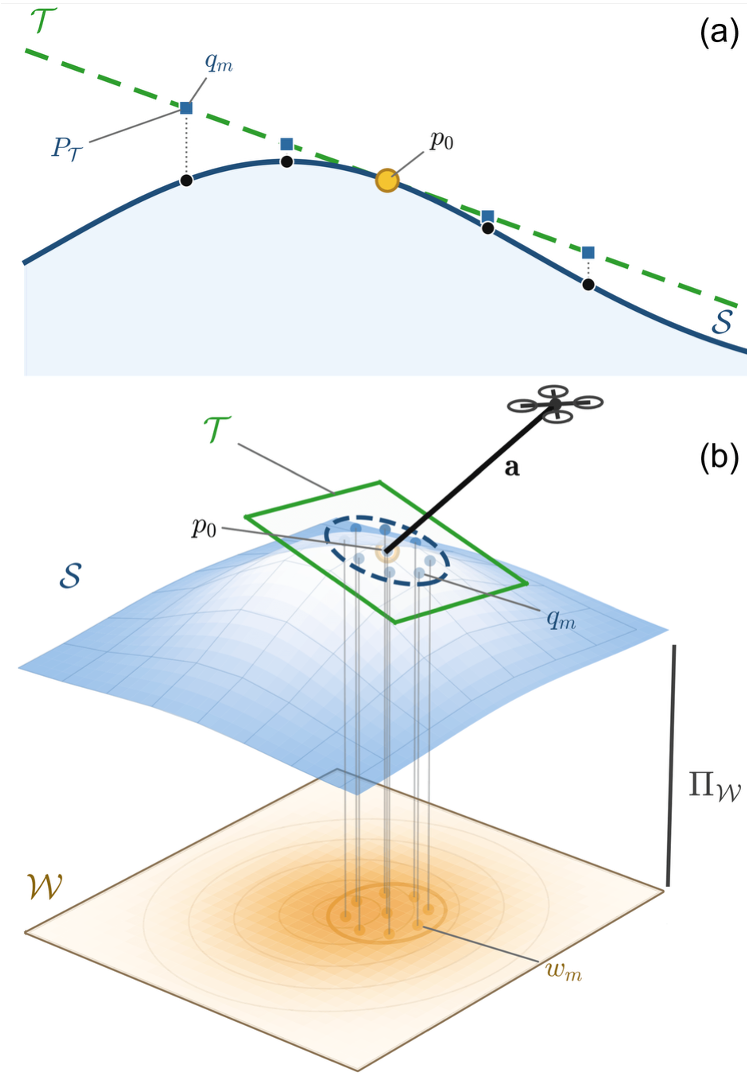}
    \caption{Illustration of concepts and maps.
    The surface $\mathcal{S}$ (a 2D manifold) is locally approximated by the tangent plane $\mathcal{T}$ at the intersection point of the sensor optical axis $\mathbf{a}$ and the surface $\mathcal{S}$.
    Within the FOV cone, a set of sampled footprint points $q_m \in \mathcal{T}$ are computed.
    The workspace mapping $\Pi_{\mathcal W}$ maps any footprint point $q_m$ to a point $w_m \in \mathcal W$ on the information map $\phi$, where the information density $\phi(w_m)$ can be evaluated. Note that although the terrain to be covered by the robot is a 2D manifold which is non-flat, the underlying information map is still 2D and can be represented by a 2D flat information map due to the map $\Pi_\mathcal{W}$.}
    \label{fig:g2s}
\end{figure}

We begin with an overview of our approach for general surfaces (Fig.~\ref{fig:g2s}).
Let $\mathcal{S}$ denote a 2D manifold embedded in 3D, and let $\mathcal{X}$ denote the full pose state of the robot including both position and orientation.
Let $a$ denote a ray corresponding to the optical axis of the FOV cone of the robot sensor, and let $p_0$ denote the first intersection point of this optical axis with the surface $\mathcal{S}$.
Let $\mathcal{T}$ denote the tangent plane of $\mathcal{S}$ at point $p_0$.
Within the FOV cone, a set of rays are sampled (indexed by $m=1,2,\cdots,M$) and the $m$-th ray has an intersection point $\mathbf q_m \in \mathcal T$, which we referred to as a footprint point.
This work always uses local tangent planes to approximate the surface during the motion of the robot, correspondingly, the footprint point on the tangent plane is also regarded as approximately located on the surface $\mathbf q_m \in \mathcal S$.

Besides, let $\Pi_{\mathcal W} : \mathcal S \rightarrow \mathcal W$ denote a map from any point on the surface to the underlying information map $\phi$ embedded in the workspace $\mathcal{W}$.
In other words, any footprint point $\mathbf q_m \in \mathcal{S}$ on the surface can be mapped to a workspace point on the information map $w_m = \Pi_{\mathcal W}(\mathbf q_m)$.
The associated Jacobian is $\mathbf J_{\mathcal W}(\mathbf q_m)=\frac{\partial \Pi_{\mathcal W}(\mathbf q_m)}
{\partial \mathbf q_m}$. \label{Define_JW}
By the chain rule, the Jacobian of a footprint sample with respect to the robot state $x$ is
\begin{equation}
\frac{\partial w_m(x)}{\partial x}
=\mathbf J_{\mathcal W}(\mathbf q_m)\frac{\partial \mathbf q_m}{\partial \boldsymbol{\mathcal X}}\frac{\partial \boldsymbol{\mathcal X}}{\partial x}.
\label{eq:general_chain_rule}
\end{equation}
The rest of this section details these concepts, derives the analytical form of the terms in Eq.~\ref{eq:general_chain_rule}, and show how the derived results are leveraged in our \abbrMethod framework.

\subsection{2D Manifold, View Cone and Tangent Planes}

\subsubsection{2D Manifold and Tangent Planes}
We represent the 2D surfaces to be covered implicitly as a manifold $\mathcal{S} = \{\mathbf{x} \in \mathbb{R}^3 \mid f(\mathbf{x}) = 0\}$, where $f: \mathbb{R}^3 \to \mathbb{R}$ is a smooth function with $\nabla f \neq \mathbf{0}$ on $\mathcal{S}$.
Let $\mathbf{p}_0$ denote the intersection point of the optical axis and the surface:
\begin{definition}[Intersection Point]\label{def:principal_contact}
The intersection point is the first intersection of the optical axis with the surface:
\begin{equation}
\mathbf{p}_0 = \mathbf{p} + l_0 \mathbf{a}, \qquad
l_0 = \min_{l > 0} \left\{ l \mid f(\mathbf{p} + l\mathbf{a}) = 0 \right\}.
\end{equation}
\end{definition}
Then, the local tangent plane at point $\mathbf{p}_0$ is:
\begin{definition}[Tangent Plane]\label{def:tangent_plane}
Let the outward unit normal at $\mathbf{p}_0$ be
\begin{equation}
\mathbf{n} = \frac{\nabla f(\mathbf{p}_0)}{\|\nabla f(\mathbf{p}_0)\|}.
\end{equation}
The tangent plane at $\mathbf{p}_0$ is
\begin{equation}
\mathcal{T} = \left\{ \mathbf{x} \in \mathbb{R}^3 \mid \mathbf{n}^\top (\mathbf{x} - \mathbf{p}_0) = 0 \right\}.
\end{equation}
\end{definition}

\subsubsection{View Cone}
When the robot position is at point $\mathbf{p}$ in the world frame, the Robot's FOV is described by a view cone.
\begin{definition}[View Cone]\label{def:view_cone}
Let $\mathbf{a} \in \mathbb{S}^2$ denote the unit vector along the optical axis, and $\theta_c \in (0, \pi/2)$ denote the cone's half-angle. The rays within the cone that point outwards are
\begin{equation}
\mathcal{D}(\theta_c) = \left\{ \mathbf{d} \in \mathbb{S}^2 \mid \mathbf{d}^\top \mathbf{a} \geq \cos\theta_c \right\}.
\end{equation}
The view cone emanating from the robot position $\mathbf{p} \in \mathbb{R}^3$ is then defined as
\begin{equation}
\mathcal{C} = \left\{ \mathbf{r} \in \mathbb{R}^3 \mid \exists\, l > 0,\; \exists\, \mathbf{d} \in \mathcal{D}(\theta_c) : \mathbf{r} = \mathbf{p} + l\mathbf{d} \right\}.
\end{equation}
\end{definition}

\subsubsection{Local Tangent Frame}

The intersection point $\mathbf{p}_0$ is the intersection of the optical axis of the sensor with the ground (information map).
The point $\mathbf{p}_0$ is the ``center'' of the sensor footprint on the local tangent plane (Fig.~\ref{fig:coordinate_schematic}).
In the \emph{perpendicular case} considered in Sec.~\ref{sec:fem}, $\mathbf{p}_0$ is same as the projected position $f_q(x)$.

Let $\mathbf{n}$ denote the normal vector of the local tangent plane.
\begin{definition}[Local Tangent Frame]\label{def:local_frame}
To construct a local coordinate frame on the tangent plane $\mathcal{T}$, we first project the robot position onto the tangent plane:
\begin{equation}
\mathbf{p}_{\perp} = \mathbf{p} - \bigl( \mathbf{n}^\top (\mathbf{p} - \mathbf{p}_0) \bigr) \mathbf{n}.
\end{equation}
We then define the local tangent frame $\mathcal{L}_T = \{\mathbf{u}, \mathbf{v}, \mathbf{n}\}$ as an orthonormal basis with origin at $\mathbf{p}_0$:
\begin{equation}
\mathbf{u} = \frac{\mathbf{p}_{\perp} - \mathbf{p}_0}{\|\mathbf{p}_{\perp} - \mathbf{p}_0\|}, \qquad
\mathbf{v} = \mathbf{n} \times \mathbf{u}, \qquad
\mathbf{n} = \frac{\nabla f(\mathbf{p}_0)}{\|\nabla f(\mathbf{p}_0)\|}.
\end{equation}
\end{definition}

Note that when the optical axis is perpendicular to the tangent plane, the projection satisfies $\mathbf{p}_{\perp} = \mathbf{p}_0$, resulting in an undefined frame $\mathcal{L}_T$.
This special case corresponds to the circular footprint model that is already discussed in the previous Sec.~\ref{sec:fem}.
The subsequent derivations assume the non-perpendicular case ($\|\mathbf{p}_{\perp} - \mathbf{p}_0\| > 0$).

These basis vectors $\{\mathbf{u}, \mathbf{v}, \mathbf{n}\}$ form a rotation matrix $\mathbf{R}_{\mathcal{T}} = [\mathbf{u}, \mathbf{v}, \mathbf{n}] \in \mathrm{SO}(3)$ that describes the orientation of this local tangent frame with respect to the world frame.
Together with the origin $\mathbf{p}_0$, the pair $(\mathbf{R}_{\mathcal{T}}, \mathbf{p}_0) \in \mathrm{SE}(3)$ defines the transformation of the local tangent frame with respect to the world frame.
Besides, a 3D point $\mathbf{q} \in \mathbb{R}^3$ on the tangent plane represented in the world frame can be transformed into its coordinates $(u, v)$ on the tangent plane with respect to the local tangent frame via a projection matrix \label{sec: LocalPuv}
\begin{equation}\label{eq:projection_matrix}
\mathbf{P}_{\mathcal{T}} = \begin{bmatrix} \mathbf{u}^\top \\ \mathbf{v}^\top \end{bmatrix} \in \mathbb{R}^{2 \times 3}, \qquad
\begin{pmatrix} u \\ v \end{pmatrix} = \mathbf{P}_{\mathcal{T}} (\mathbf{q} - \mathbf{p}_0).
\end{equation}
In other words, the matrix $\mathbf{P}_{\mathcal{T}}$ extracts the local coordinates of a 3D point on the tangent plane.

\begin{definition}[Geometric Parameters]\label{def:geometric_params}
We define the following geometric quantities to help the derivation later: the height $h$, the horizontal offset $\rho$, and the incidence angle $\theta_{\mathrm{inc}}$:
\begin{equation}
h = \mathbf{n}^\top (\mathbf{p} - \mathbf{p}_0), \;
\rho = \|\mathbf{p}_{\perp} - \mathbf{p}_0\|, \;
\theta_{\mathrm{inc}} = \arctan\!\left( \frac{\rho}{h} \right).
\end{equation}
These quantities satisfy $\mathbf{p} - \mathbf{p}_0 = \rho\mathbf{u} + h\mathbf{n}$ and $\rho = h \tan\theta_{\mathrm{inc}}$.
\end{definition}
If $\mathbf{p}_0 = \mathbf{p} + l_0 \mathbf{a}$ with $l_0 > 0$ and $\mathbf{n}$ points outward from the surface, then the incidence angle satisfies
\begin{equation}\label{eq:incidence_angle}
\cos\theta_{\mathrm{inc}} = -\mathbf{n}^\top \mathbf{a}, \qquad
\sin\theta_{\mathrm{inc}} = \sqrt{1 - (\mathbf{n}^\top \mathbf{a})^2}.
\end{equation}
Intuitively, $h$ is the height of the robot above the tangent plane, $\rho$ is the horizontal offset from $\mathbf{p}_0$ to the robot's projection $\mathbf{p}_\perp$, and $\theta_{\mathrm{inc}}$ measures the incidence angle between the optical axis and the surface normal (Fig.~\ref{fig:coordinate_schematic}).
Larger values of $\theta_{\mathrm{inc}}$ correspond to more oblique viewing angles, and when $\theta_{\mathrm{inc}} = 0$, the optical axis is perpendicular to the tangent plane.

\subsection{Sensor Footprint Geometry for Oblique Views}

We now investigate the geometry of the sensor footprint on the tangent plane.

\begin{definition}[Local Sensor Frame]\label{def:cone_basis}
As aforementioned, assuming $\sin\theta_{\mathrm{inc}} = \|\mathbf{a} \times \mathbf{n}\| > 0$ (i.e., the non-perpendicular case), we define a local frame attached to the sensor by orthonormal vectors $\mathcal{L}_S=\{\mathbf{a}, \mathbf{e}_1, \mathbf{e}_2\}$, where
\begin{equation}
\mathbf{e}_2 = -\frac{\mathbf{a} \times \mathbf{n}}{\|\mathbf{a} \times \mathbf{n}\|}, \qquad
\mathbf{e}_1 = \mathbf{e}_2 \times \mathbf{a}.
\end{equation}
When expanded in the local tangent frame basis $\{\mathbf{u}, \mathbf{v}, \mathbf{n}\}$, these vectors take the form
\begin{equation}\label{eq:cone_basis_local}
\begin{aligned}
\mathbf{a} &= -\sin\theta_{\mathrm{inc}}\, \mathbf{u} - \cos\theta_{\mathrm{inc}}\, \mathbf{n}, \\
\mathbf{e}_1 &= \cos\theta_{\mathrm{inc}}\, \mathbf{u} - \sin\theta_{\mathrm{inc}}\, \mathbf{n}, \\
\mathbf{e}_2 &= -\mathbf{v}.
\end{aligned}
\end{equation}
We refer to this frame $\mathcal{L}_S$ as the local sensor frame.
\end{definition}

Within the local sensor frame $\mathcal{L}_S$, any ray within the cone can be described by two angles $(\beta,\varphi)$ with respect to the optical axis:
For any angle $\beta \in [0, \theta_c]$ from the optical axis and azimuth $\varphi \in [0, 2\pi)$, the corresponding ray direction within the cone can be represented in the world frame as:
\begin{equation}\label{eq:ray_direction}
\mathbf{d}(\beta, \varphi) = \cos\beta\, \mathbf{a} + \sin\beta \cos\varphi\, \mathbf{e}_1 + \sin\beta \sin\varphi\, \mathbf{e}_2 .
\end{equation}
Define an auxiliary variable $D(\beta, \varphi)$ as
\begin{equation}\label{eq:D_function}
D(\beta, \varphi) = \cos\theta_{\mathrm{inc}} \cos\beta + \sin\theta_{\mathrm{inc}} \sin\beta \cos\varphi.
\end{equation}
Note that $D(\beta, \varphi) = -\mathbf{n}^\top \mathbf{d}(\beta, \varphi)$, which means:
\begin{itemize}
    \item $D < 0$: the ray points away from the tangent plane;
    \item $D = 0$: the ray is parallel to the tangent plane;
    \item $D > 0$: the ray points towards the tangent plane.
\end{itemize}
We only need to consider $D > 0$ in this paper, which is guaranteed for all rays within the view cone when $\theta_{\mathrm{inc}} + \theta_c < \pi/2$.

\begin{definition}[Sensor Footprint Geometry]\label{lem:footprint_general}
For any ray $(\beta,\varphi)$ within the view cone, 
let
\begin{align}\label{Q(b,p)}
  \mathbf q(\beta,\varphi)=\mathbf p+l^*(\beta,\varphi)\mathbf d(\beta,\varphi)  
\end{align}
denote the corresponding footprint point, i.e., the intersection of the ray
with the tangent plane $\mathcal T$ expressed in the world frame, where
\begin{align}\label{eq:intersection_distance}
l^*(\beta,\varphi)=\frac{h}{D(\beta,\varphi)}.
\end{align}
The footprint boundary can be obtained by setting $\beta = \theta_c$ and varying $\varphi \in [0, 2\pi)$.
\end{definition}
As a special case, when $\theta_{\mathrm{inc}} = 0$ (i.e., the optical axis is perpendicular to the tangent plane $\mathcal{T}$), we have $D(\beta, \varphi) = \cos\beta$, and the footprint point $\mathbf q(\beta,\varphi)$ reduces to a circular footprint centered at $\mathbf{p}_0$ with radius $h \tan\theta_c$ as aforementioned in Sec.~\ref{sec:fem}.

The footprint geometry in Def.~\ref{lem:footprint_general} provides a way to describe the sensor footprint distribution $\gamma(w, x)$ (Sec.~\ref{sec:fem}): Given a point $w \in \mathcal{W}$ on the tangent plane, the value $\gamma(w, x)$ can be computed by checking whether $w$ lies within the boundary of the footprint on the tangent plane.

\subsection{Pose Jacobians}

To incorporate the derivation into gradient-based trajectory optimization, we need to compute the gradient of the Lagrangian with respect to the robot state, as shown in Eq.~\eqref{meth:LD}.
In particular, the term $\frac{\partial \mathcal{E}'_\gamma(\phi, x)}{\partial x}$ in Eq.~\eqref{eq:a_esdf_gamma} requires the Jacobian $\frac{\partial w_m(x)}{\partial x}$ for each sampled point within the footprint.
We derive the Jacobians of these sampled points with respect to the robot pose, and examine whether these Jacobians can be reused across optimization iterations.
We derive the Jacobians by fixing the local tangent frame at each time instant; that is, we treat $\mathbf{p}_0$ and $\mathbf{n}$ as constants in the derivation.

\begin{definition}[Robot Pose State]\label{def:pose_state}
Let $\boldsymbol{\mathcal{X}} = [\mathbf{p}^\top, \boldsymbol{\Theta}^\top]^\top \in \mathbb{R}^6$ denote the full pose state, where $\mathbf{p} \in \mathbb{R}^3$ is the position and $\boldsymbol{\Theta} = [\phi_r, \theta_p, \psi_y]^\top$ are the roll, pitch, and yaw Euler angles, respectively.
\end{definition}

Note that the pose state $\boldsymbol{\mathcal{X}}$ is a subvector of the full robot state $x$, and its Jacobian with respect to $x$, $\frac{\partial\boldsymbol{\mathcal{X}}}{\partial x}$ is therefore a constant matrix that selects the corresponding entries from $x$. 

To simplify the derivation, we set the optical axis $\mathbf{a}$ to align with the $x$-axis of the local sensor frame $\mathcal{L}_S$.\footnote{If not, one can re-define the local sensor frame so that they align.}
Note that, the direction of the optical axis in the world frame is determined by pitch and yaw only.
In other words, the roll angle does not affect the orientation and shape of the view cone.

For the rest of this section, we list the key results to show the idea, and provide the detailed proofs in Appendix.

\begin{lemma}[Position Jacobian]\label{lem:position_jacobian}
For any ray $(\beta, \varphi)$ with $D(\beta, \varphi) > 0$, let $\mathbf{q}(\beta, \varphi) = \mathbf{p} + l^* \mathbf{d}$ denote the corresponding \emph{footprint point}, i.e., the point where the ray intersects with the tangent plane, represented in the world frame.
The Jacobian of $\mathbf{q}$ with respect to the robot position $\mathbf{p}$ is
\begin{equation}\label{eq:position_jacobian}
\mathbf{J}_p(\beta, \varphi) = \frac{\partial \mathbf{q}}{\partial \mathbf{p}} = \mathbf{I}_3 + \frac{1}{D(\beta, \varphi)}\, \mathbf{d}(\beta, \varphi)\, \mathbf{n}^\top \in \mathbb{R}^{3 \times 3}.
\end{equation}
\end{lemma}

\begin{remark}\label{remark:linearity}
This lemma indicates that, for a fixed tangent plane (i.e., both $(\beta,\varphi)$ and $\mathbf{n}$ are constant), the position Jacobian $\mathbf{J}_p$ is then a constant matrix which suggests the linearity between the sensor footprint and the robot position: when the robot translates, the footprint size grows larger or smaller proportionally with respect to the translation.
Note that, this fixed tangent can have any orientation and does not have to be horizontal.
When the tangent plane varies, (i.e., $(\beta,\varphi)$ and $\mathbf{n}$ are varying), the position Jacobian $\mathbf{J}_p$ is no longer a constant matrix: the first term $I_3$ remains constant while the second term $\frac{1}{D(\beta, \varphi)}\, \mathbf{d}(\beta, \varphi)\, \mathbf{n}^\top \in \mathbb{R}^{3 \times 3}$ depends on the orientation of the tangent plane.
In practice, to compute $\mathbf{J}_p$, we only need to evaluate the second term there.
\end{remark}

For the orientation component, we first derive the Jacobian of $\mathbf{q}$ with respect to the optical axis $\mathbf{a}$, then relate $\mathbf{a}$ to the Euler angles $\boldsymbol{\Theta}$ via the chain rule.
Derivation is in the Appendix.
\begin{lemma}[Orientation Jacobian]\label{lem:orientation_jacobian}
For any ray $(\beta, \varphi)$ with $D(\beta, \varphi) > 0$, the Jacobian of the footprint point $\mathbf{q}$  with respect to the optical axis of the robot is
\begin{equation}\label{eq:orientation_jacobian}
\mathbf{J}_a(\beta, \varphi) = \frac{\partial \mathbf{q}}{\partial \mathbf{a}} = \frac{h}{D}\, \frac{\partial \mathbf{d}}{\partial \mathbf{a}} - \frac{h}{D^2}\, \mathbf{d} \left( \frac{\partial D}{\partial \mathbf{a}} \right)^\top \in \mathbb{R}^{3 \times 3},
\end{equation}
where
\begin{align}\label{jacobian}
    &\frac{\partial \mathbf{d}}{\partial \mathbf{a}} = \cos\beta\, \mathbf{I}_3 + \sin\beta \cos\varphi\, \frac{\partial \mathbf{e}_1}{\partial \mathbf{a}} + \sin\beta \sin\varphi\, \frac{\partial \mathbf{e}_2}{\partial \mathbf{a}}, \\ 
    &\frac{\partial D}{\partial \mathbf{a}} =\frac{-\sin\theta_{\mathrm{inc}} \cos\beta + \cos\theta_{\mathrm{inc}} \sin\beta \cos\varphi}{\sin\theta_{\mathrm{inc}}} \mathbf{n}.
\end{align}
\end{lemma}

To relate the optical axis $\mathbf{a}$ to the robot's orientation, we introduce the Jacobian $\mathbf{J}_{a/\Theta} = \frac{\partial \mathbf{a}}{\partial \boldsymbol{\Theta}}$, which we refer to as the axis-to-Euler Jacobian.
\begin{lemma}[Axis-to-Euler Jacobian]\label{lem:axis_euler}
The optical axis in the world frame is
\begin{equation}\label{eq:optical_axis}
\mathbf{a} = \begin{bmatrix}
\cos\theta_p \cos\psi_y \\[2pt]
\cos\theta_p \sin\psi_y \\[2pt]
-\sin\theta_p
\end{bmatrix}
\end{equation}
and $\mathbf{J}_{a/\Theta}$ is
\begin{equation}\label{eq:axis_euler_jacobian}
\mathbf{J}_{a/\Theta} = \frac{\partial \mathbf{a}}{\partial \boldsymbol{\Theta}} = \begin{bmatrix}
0 & -\sin\theta_p \cos\psi_y & -\cos\theta_p \sin\psi_y \\[2pt]
0 & -\sin\theta_p \sin\psi_y & \cos\theta_p \cos\psi_y \\[2pt]
0 & -\cos\theta_p & 0
\end{bmatrix}.
\end{equation}
\end{lemma}

Combining the position Jacobian (Lemma~\ref{lem:position_jacobian}) and the orientation Jacobian (Lemmas~\ref{lem:orientation_jacobian}--\ref{lem:axis_euler}), we obtain the Jacobian with respect to the full pose.

\begin{lemma}[Full Pose Jacobian]\label{lem:pose_jacobian}
For a ray $(\beta, \varphi)$, the Jacobian of a footprint point $\mathbf{q}$ with respect to the full pose state $\boldsymbol{\mathcal{X}}$ is
\begin{equation}\label{eq:full_pose_jacobian}
\frac{\partial \mathbf{q}}{\partial \boldsymbol{\mathcal{X}}} = \begin{bmatrix}
\mathbf{J}_p(\beta, \varphi) & \mathbf{J}_a(\beta, \varphi)\, \mathbf{J}_{a/\Theta}
\end{bmatrix} \in \mathbb{R}^{3 \times 6}.
\end{equation}
\end{lemma}

The above Jacobians describe how a footprint point moves with respect to the robot pose.

\subsection{Extract Information}
\label{subsec:workspace_projection}
We now discuss the map $\Pi_{\mathcal W}$ that maps from any point on the tangent plane to the underlying information map $\phi$ embedded in the workspace $\mathcal{W}$.
Depending on the representation of the information map, $\Pi_{\mathcal W}$ takes different forms.
This work considers the following two cases.

\subsubsection{Case I: The information map $\phi$ aligns with the manifold $\mathcal{S}$} \label{case1}
When the information distribution is defined on the 2D manifold 
$\mathcal{S}$, the map $\Pi_\mathcal{W}$ is an identity map and the point on the information map is same as the point on the manifold $w_m = \mathbf{q}_m$.
Correspondingly, $\mathbf J_{\mathcal W}$ is an identity matrix:
\begin{align}
    \mathbf J_{\mathcal W} = \begin{bmatrix}
        1 & 0 & 0\\
        0 & 1 & 0\\
        0 & 0 & 1
    \end{bmatrix}.
\end{align}
In practice, the information map on a manifold can be represented in different ways, such as using a mesh~\cite{dong2025ergodic}.
This work uses a point cloud to represent the information map, which samples a finite number of points $\mathcal{W}_S$ from the manifold and assigns each point $w_s \in \mathcal{W}_S$ a probability value while ensuring that all these values sum to one. When computing $\phi(w)$ for any point $w \in \mathbb{W}$, we find the nearest sampled point in $w_s \in \mathcal{W}_S$ and use the value $\phi_(w_s)$ for $\phi(w)$.

\subsubsection{Case II: The information map $\phi$ is defined over a flat 2D plane} \label{case2}
When the surface $\mathcal{S}$ is relatively simple, the information map related to $\mathcal{S}$ can be possibly described by a 2D flat information map.
In this case, $\Pi_\mathcal{W}$ projects each sampled footprint point $\mathbf{q}_m$ to a point $w_m$ on the 2D information map.
Correspondingly, $\mathbf J_{\mathcal W}$ is a constant matrix that describes the projection.
For instance, when $\phi$ is defined over the $xy$ flat plane, the map $\Pi_\mathcal{W}$ extracts the $(x,y)$ components from $\mathbf q_m \in \mathbb{R}^3$, and 
\begin{align}
    \mathbf J_{\mathcal W} = \begin{bmatrix}
        1 & 0 & 0\\
        0 & 1 & 0
    \end{bmatrix}.
\end{align}

\subsection{Jacobians and Computational Reuse}\label{sec:3d_jac_reuse}

Now we have all we need to obtain the analytical form of the Jacobian in Eq.~\ref{eq:general_chain_rule}, which is stated in the following theorem.

\begin{theorem}[Footprint Point Jacobian]\label{thm:sampled_jacobian}
For each sampled footprint point $q_m$ with parameters $(\beta_m,\varphi_m)$ defined in
Definition \ref{lem:footprint_general}, as overviewed in Sec. \ref{sec:workspace_mapping}, the Jacobian of a footprint point with respect to the robot state $x$ is:
\begin{equation}
\frac{\partial w_m(x)}{\partial x}
=
\mathbf J_{\mathcal W}(\mathbf q_m)
\frac{\partial \mathbf q_m}{\partial \boldsymbol{\mathcal X}}
\frac{\partial \boldsymbol{\mathcal X}}{\partial x},
\end{equation}
where $\mathbf J_{\mathcal W}(\mathbf q_m)$ is presented in Sec.~\ref{subsec:workspace_projection}, $\frac{\partial \mathbf q_m}{\partial \boldsymbol{\mathcal X}}$ is the full pose Jacobian as derived in Lemma~\ref{lem:pose_jacobian}, and $\frac{\partial \boldsymbol{\mathcal X}}{\partial x}$ is a matrix in $\mathbb{R}^{6\times n}$ that extracts the corresponding position and orientation components from the robot state vector $x$.
\end{theorem}

While the entire Jacobian in Theorem~\ref{thm:sampled_jacobian} is pose-dependent, some components in its computation can be reused. This computational saving arises in several parts.
First, as discussed in Remark \ref{remark:linearity}, the evaluation of $\mathbf{J}_p(\beta, \varphi)$ requires only computing the second term that is pose dependent.
Second, the matrices $\mathbf J_{a/\Theta}$ and $\frac{\partial \boldsymbol{\mathcal X}}{\partial x}$ do not depend on the footprint index $m$.
They are therefore computed once (Line~1 in Alg.~\ref{alg:al_ilqr}) and reused for all sampled footprint points during gradient computation (Line~6). 
Third, for any sampled ray with parameters $\{(\beta_m,\varphi_m)\}_{m=1}^M$, these parameters are fixed, and the associated trigonometric terms $\{\cos\beta_m, \sin\beta_m\cos\varphi_m, \sin\beta_m\sin\varphi_m\}$ remain constant during optimization, which can be computed once and reused.
Finally, in both cases described in Sec.~\ref{subsec:workspace_projection}, the Jacobian $\mathbf J_{\mathcal W}$ reduces to a constant matrix, which can be computed once (Line~1 in Alg.~\ref{alg:al_ilqr}) and reused across all iterations.

\begin{remark}
The Jacobians become singular when 
$D(\beta,\varphi)\to 0$, i.e., when a sampled ray $\mathbf{d}(\beta,\varphi)$ becomes parallel to the tangent plane ($\mathbf{n}^\top \mathbf{d} \to 0$).
This singularity can be avoided by skipping this sampled ray since it is parallel to and has no intersection point with the tangent plane.
In addition, the orientation Jacobian Eq.~\ref{eq:orientation_jacobian} contains the factor $1/\sin\theta_{\mathrm{inc}}$ and therefore becomes singular when $\sin\theta_{\mathrm{inc}}\to 0$, i.e., when the optical axis $\mathbf{a}$ is perpendicular to the tangent plane.
As aforementioned, this case degenerates to the simpler case presented in Sec.~\ref{sec: NA}.
\end{remark}

\subsection{Local Tangent Plane Approximation Errors}
\label{sec:tangent_error}

Under the Assumption \ref{assume1}, we bound the resulting approximation error and its effect on the footprint ergodic metric $\mathcal{E}_\gamma$.
\begin{figure}[tb]
    \centering
    \includegraphics[width=\linewidth]{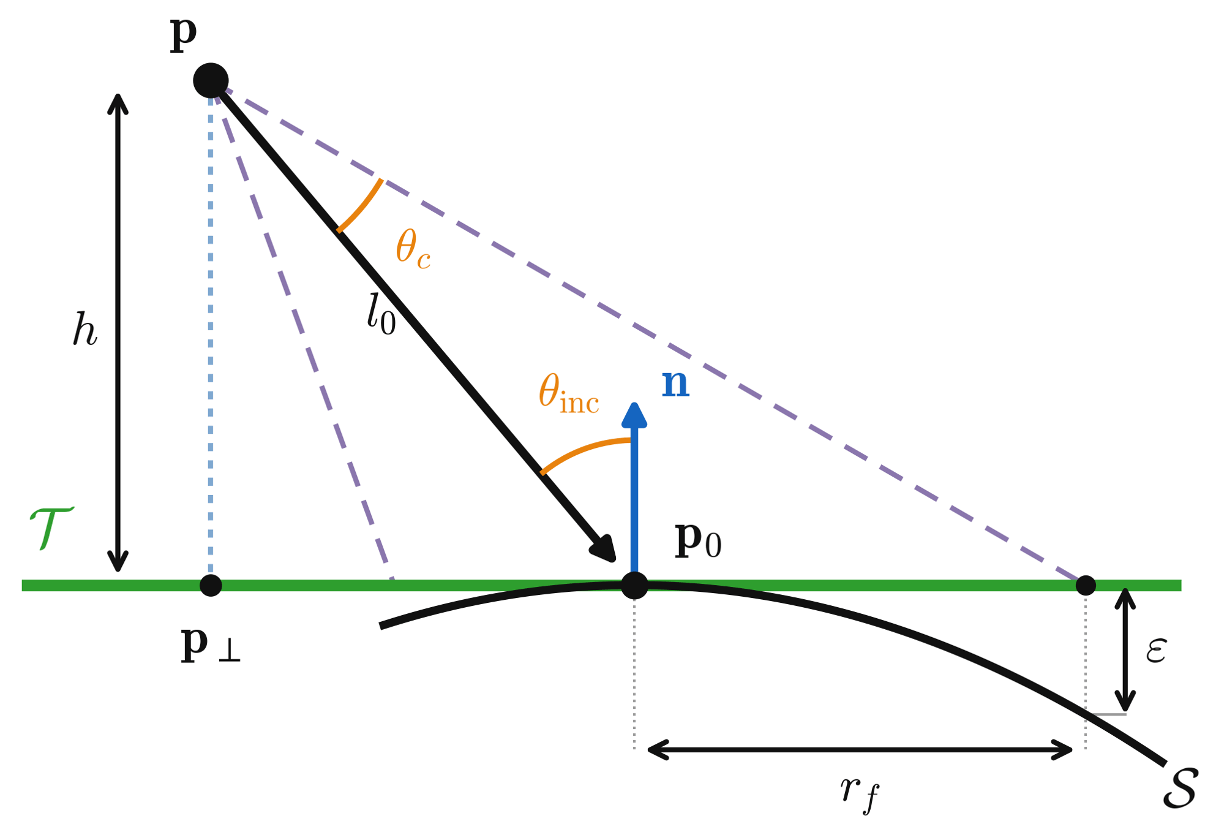}
    \vspace{-5mm}
     \caption{Geometry of the local tangent-plane approximation error. The robot at $\mathbf{p}$
    views the surface $\mathcal{S}$ along its optical axis, which reaches the contact point
    $\mathbf{p}_0$ at distance $l_0$; the view cone (dashed) has half-angle $\theta_c$.
    $\mathcal{T}$ is the tangent plane of $\mathcal{S}$ at $\mathbf{p}_0$, $\mathbf{n}$ its outward
    normal, and $\theta_{\mathrm{inc}}$ the incidence angle between the optical axis and
    $\mathbf{n}$.  $\mathbf{p}_{\perp}$ is the projection of $\mathbf{p}$ onto $\mathcal{T}$ and
    $h=l_0\cos\theta_{\mathrm{inc}}$ the height of $\mathbf{p}$ above it. The footprint radius
    $r_f$ is the largest in-plane distance of a footprint point from $\mathbf{p}_0$, and
    $\varepsilon$ is the deviation of $\mathcal{T}$ from $\mathcal{S}$ at that point, bounded by
    $\varepsilon \le \tfrac12 \kappa_{\max} r_f^{2}$ (Theorem~\ref{thm:tangent_error}). 
    This figure helps illustrate the derivation of approximation errors.}
    \label{fig:error_analysis}
\end{figure}

\subsubsection{Geometric Error}

As illustrated in Fig.~\ref{fig:error_analysis}, consider an optical axis of the sensor intersecting a surface $\mathcal{S}$ at point $p_0$ with $\mathcal{T}$ denoting the local tangent plane at $p_0$.
Let $\kappa_{\max}$ denote the largest absolute value of the principal curvatures of $\mathcal{S}$ within the FOV, which is finite due to Assumption~\ref{assume1}. 

For any footprint point $p$ within the FOV in the local tangent plane, let $q$ denote the corresponding point on the surface and let $r$ denote the in-plane distance of that footprint point $p$ from $p_0$. 
Based on differential geometry, with a second-order Taylor expansion of the surface $\mathcal{S}$ at $p_0$, the distance $\varepsilon(r)$ of $q$ from the point $p$ in the tangent plane $\mathcal{T}$ is bounded by
\begin{equation}
\varepsilon(r)\le\tfrac12\,\kappa_{\max}\,r^2+O(r^3).
\label{eq:eps_geo}
\end{equation}
\begin{theorem}[Approximation error]\label{thm:tangent_error}
Let $r_f$ be the footprint radius, i.e., the largest $r$ within the FOV. Under Assumption~\ref{assume1}, every footprint point on $\mathcal{T}$ deviates from
$\mathcal{S}$ by at most
\begin{align}
\varepsilon&\le\tfrac12\,\kappa_{\max}\,r_f^2,\\
r_f&=l_0\sin\theta_c/\cos(\theta_{\mathrm{inc}}+\theta_c).
\label{eq:eps_full}
\end{align}
\end{theorem}

\begin{proof}
Note that, $r_f$ is finite for $\theta_{\mathrm{inc}}+\theta_c<\pi/2$.
Ignoring the higher-order term in~\eqref{eq:eps_geo}, $\varepsilon(r)\le\tfrac12\kappa_{\max}r^2$, so
it suffices to bound $r$ by the footprint radius $r_f$. Let $\mathbf{p}_\perp$ be the projection of $\mathbf{p}$ onto $\mathcal{T}$ (Def.~\ref{def:local_frame}), at height $h=\mathbf{n}^\top(\mathbf{p}-\mathbf{p}_0)=l_0\cos\theta_{\mathrm{inc}}$. 
The distance between $\mathbf{p}_\perp$ and $\mathbf{p}_0$ is $\rho=h\tan\theta_{\mathrm{inc}}$.
Following the frame defined in Def.~\ref{def:cone_basis}, in the plane containing $\mathbf{a}$ and $\mathbf{n}$, the angle between the outermost cone ray ($\beta=\theta_c$, $\varphi=\pi$) and the line between $\mathbf{p}$ and $\mathbf{p}_\perp$ is $\theta_{\mathrm{inc}}+\theta_c$ with $D(\theta_c,\pi)=
\cos(\theta_{\mathrm{inc}}+\theta_c)$ by~\eqref{eq:D_function}.
The outermost cone ray meets $\mathcal{T}$ at a distance of $h\tan(\theta_{\mathrm{inc}}+\theta_c)$ from $\mathbf{p}_\perp$.
With $h=l_0\cos\theta_{\mathrm{inc}}$:
\begin{align}
r_f &= h\tan(\theta_{\mathrm{inc}}+\theta_c)-h\tan\theta_{\mathrm{inc}} \notag\\
    &= \frac{h\sin\theta_c}{\cos(\theta_{\mathrm{inc}}+\theta_c)\cos\theta_{\mathrm{inc}}}
     = \frac{l_0\sin\theta_c}{\cos(\theta_{\mathrm{inc}}+\theta_c)},
\label{eq:rf_proof}
\end{align}
where the second equality uses
$\tan(\theta_{\mathrm{inc}}+\theta_c)-\tan\theta_{\mathrm{inc}}=\sin\theta_c/[\cos(\theta_{\mathrm{inc}}+\theta_c)\cos\theta_{\mathrm{inc}}]$.
Substituting $r\le r_f$ into $\varepsilon(r)\le\tfrac12\kappa_{\max}r^2$ yields~\eqref{eq:eps_full}.
\end{proof}
It shows that $\varepsilon$ grows linearly with the curvature and quadratically with the footprint radius, and the maximum footprint radius grows linearly with $l_0$, the distance between $\mathbf{p}$ and $\mathbf{p}_0$.
Intuitively, it indicates that the approximation error is large when the robot is far away from the surface, or when the robot has a very oblique view of the surface.

\subsubsection{Effect on the Footprint Ergodic Metric}
The footprint ergodic metric $\mathcal{E}_\gamma$ is a nonlinear function of the trajectory, which makes it challenging to derive a closed-form representation of how $\varepsilon$ propagates to $\mathcal{E}_\gamma$.
We only provide an intuitive discussion.
The metric does not use the footprint points directly.
It uses only a finite set of low-frequency Fourier coefficients of the time-averaged statistics of the footprint trajectory (Def.~\ref{def:tas_sft}).
The geometric approximation error (Eq.~\eqref{eq:eps_full}) $\varepsilon$ perturbs each coefficient bounded by an amount proportional to $\varepsilon$, which is often a small amount and usually affects the high frequency components of the Fourier coefficients.
Since in practice $\mathcal{E}_\gamma$ is computed by using only a finite number of low-frequency modes with decaying weights $\Lambda_k$, the resulting ergodic metrics are less sensitive to such geometric approximation error.

Finally, for surfaces with large curvatures, or surfaces that violate Assumption~\ref{assume1} (e.g., 3D objects), a promising future direction is to iteratively apply our approach to a local region on such surfaces as the robot moves. Besides, similar to regular ergodic metric, the proposed footprint ergodic metric (Def.~\ref{esdf:eqn:ergodic_metric}) also suffers from distortion in the metrics, especially when the surface has large curvature.
Such distortion can be handled by using the eigenfunctions of the Laplace-Beltrami operator on the surface mesh, which form an intrinsic Fourier-like basis and reduce to the cosine modes on a flat domain~\cite{Sharp:2020:LNT, 1631196,dong2025ergodic,bilaloglu2024tactile}.
As it is not the major focus of this paper, we briefly showcase its usage in the experimental results (Sec.~\ref{3Dreal}).

\section{Experimental Results}
\label{sec:result}
\graphicspath{{figures/}}
\begin{figure*}[tb]
    \centering
    \includegraphics[width=0.95\textwidth,trim=1 1 1 1,clip]{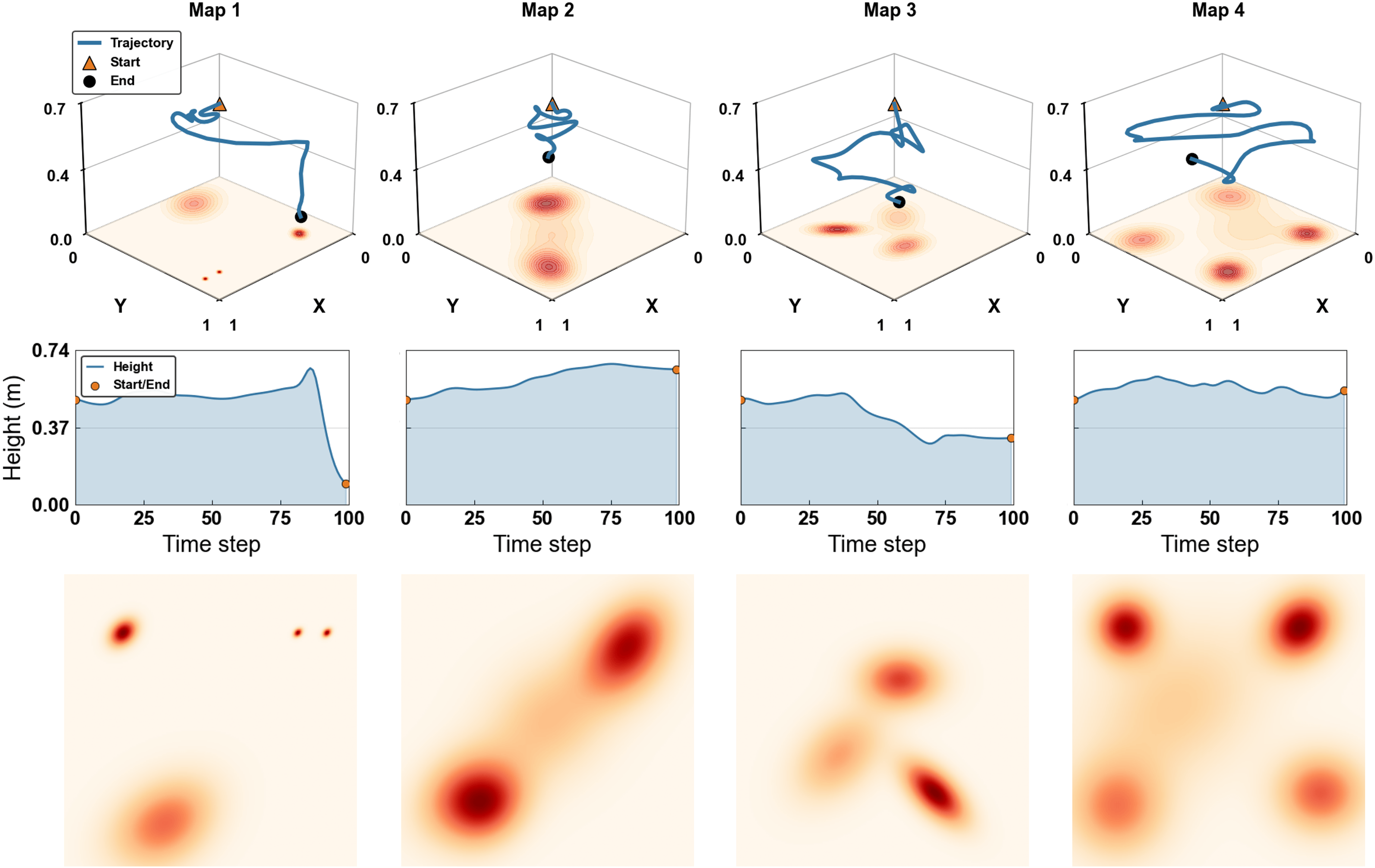}
    \caption{ The experimental results in four different information maps. The first row (I) visualizes the planned trajectories by our \abbrMethod.
 The middle row (II) shows the corresponding height value ($z$ coordinates) of the robot along the planned trajectories by our \abbrMethod. The last row (III) shows four information maps with various regions with widespread and concentrated information.
  Our \abbrMethod can adapt the flying height of the robot and the sensor footprint based on the information map.}
    \label{fig: Different Maps}
\end{figure*}
We first run experiments with 2D information maps (Sec.~\ref{REM}-\ref{CCRA}), then with 3D complex terrains and 3D Objects (Sec.~\ref{Terrains_3D}-\ref{3Dreal}), and finally with real robots (Sec.~\ref{Experiment}).

\subsection{Experimental Settings}

\subsubsection{Dynamics and Parameters}

For experiments with 2D information maps, the robot dynamics is double integrators with the robot state including $(x,y,z)$ positions and $(v_x,v_y,v_z)$ velocities and the robot control including $(a_x,a_y,a_z)$ accelerations.
The position of the robot is limited to a bounded space $[0,1]^3$ and the controls are bounded component-wise within $[-0.5, 0.5]$.
We adopt the uniform circular footprint model in Example~\ref{example1}, where the footprint radius scales with altitude as 
$r(t) = k_h f_h(x(t))$ and $k_h = 0.25$. 
The footprint ergodic metric is approximated using deterministic uniform sampling inside the unit disk with Cartesian grid resolution $\Delta = 0.35$, which corresponds to $M=25$ sampled points within the footprint.
The number of cosine Fourier basis for computing the ergodicity is  $10 \times 10$.

\subsubsection{Baseline Methods}\label{ESDF_baseline}
We consider the following baselines.

\begin{itemize}
\item Baseline 1 Regular Ergodic Metric:
It optimizes the original ergodic metric $\mathcal{E}(\phi,x)$ with Dirac delta function~\cite{mathew2011metrics}.

\item Baseline 2 Fixed Sampled Footprint: \label{fixed_base}
It optimizes the proposed footprint ergodic metric yet with a fixed footprint, which is equivalent to that the robot flies with a fixed altitude, resulting in a constant footprint radius.

\item Baseline 3 Fixed Gaussian Footprint:
Following the literature~\cite{ayvali2017ergodic,dong2025ergodic}, the uniform footprint in Eq.~\ref{eq:gam} is replaced by a Gaussian distribution $\mathcal{N}(\mu(t),\sigma)$ whose mean value $\mu(t)$ is same as the robot $x,y$ position, and its variance $\sigma$ is constant.

\item Baseline 4 HEDAC:
It is based on the heat equation driven area coverage method~\cite{bilaloglu2023whole,bilaloglu2024tactile}, which guides robot motion by the gradient of the evolving heat field instead of trajectory optimization.
We implement this approach with a fixed size sensor footprint in our setting as a baseline for comparison.

\end{itemize}

Note that, all baselines have a fixed sensor footprint, while our approach has a dynamic footprint and the trajectory is optimized in 3D space, including altitude, without orientation.
Fig.~\ref{fig: Different Maps} shows the planned trajectories by our \abbrMethod.

\subsection{Footprint Ergodic Metrics in Different Maps}\label{FEM}

We run the planners in four different maps as shown in Fig.~\ref{fig: Ergo} and compare their runtime and the corresponding ergodic metrics.
This section discusses the ergodicity, and we discuss the runtime in Sec.~\ref{CCRA} later.

\subsubsection{Against Regular Ergodic Metric}\label{REM}
\begin{figure*}[tb]
    \centering
    \includegraphics[width=\textwidth]{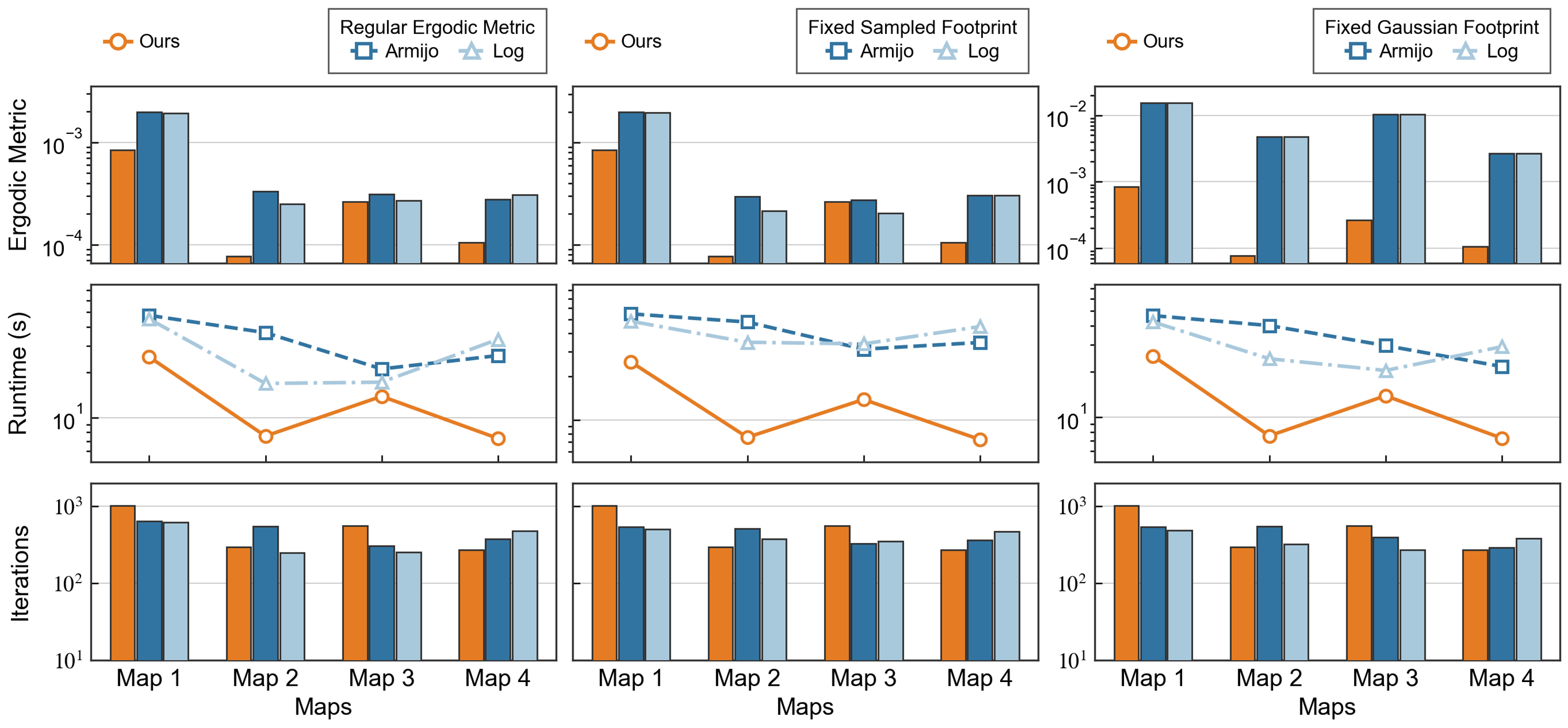}
    \vspace{-5mm}
    \caption{We conducted a comparison using three baselines to evaluate the trajectory optimization over 100 time steps across four different information maps shown in Fig~\ref{fig: Different Maps}. In the first row (a), we present the comparison of the ergodic metric. The second row (b) displays the time cost for convergence, while the third row (c) presents the number of iterations required. Columns (I), (II), and (III) correspond to the results from Baselines 1, 2, and 3, respectively.}
    \label{fig: Ergo}
\end{figure*}

The results are shown in the first column in Fig.~\ref{fig: Ergo}.
Our \abbrMethod achieves better (lower) ergodicity than the baselines with or without log-spaced search when using the regular ergodic metric.
In particular, the ergodicity is reduced by approximately $57\%$ on Map $1$, 
$76\%$ on Map $2$, $15\%$ on Map $3$, and $62\%$ on Map $4$, which verifies the benefits of considering dynamic footprint.
In addition to the improved ergodicity, our \abbrMethod consistently reduces the total runtime across all maps. 
A detailed discussion on runtime is provided in Sec.~\ref{sec:comp_analysis}.

\subsubsection{Against Fixed Sampled Footprint} \label{FFS}
The radius of the fixed footprint is selected as the one  corresponding to the midpoint of the allowable altitude range.
As shown in the middle column in Fig.~\ref{fig: Ergo}, our \abbrMethod achieves lower ergodicity than the baseline on all four maps, where the ergodicity is reduced by approximately 
$57\%$ on Map $1$, $74\%$ on Map $2$, and $66\%$ on Map $4$. 
In particular, on Map $3$, where the information distribution is relatively smooth and has a single concentrated area, both methods achieve comparable performance (with only $3.7\%$ difference).
On other maps, the differences in the ergodicity are larger.
When the information map has both concentrated and wide-spread information, a single fixed-size footprint (as in the baseline) covers the map poorly, while our \abbrMethod covers well due to the dynamic footprint.

\subsubsection{Against Fixed Gaussian Footprint} \label{GBF}

When compared to the baseline \cite{ayvali2017ergodic,dong2025ergodic} in the last column in Fig.~\ref{fig: Ergo}. 
Our \abbrMethod reduces the ergodicity by approximately $18\times$ on Map $1$, $61\times$ on Map $2$, $40\times$ on Map $3$, and $25\times$ on Map $4$.
The reason is that a Gaussian footprint with fixed covariance better covers low-frequency components in the Fourier spectrum and cannot ergodically cover both concentrated and wide-spread information simultaneously, while our \abbrMethod can adjust the footprint size when needed.

\subsubsection{Against HEDAC}\label{CaH}
\begin{figure}[tb]
    \centering
    \includegraphics[width=\linewidth,trim=2 2 2 2,clip]{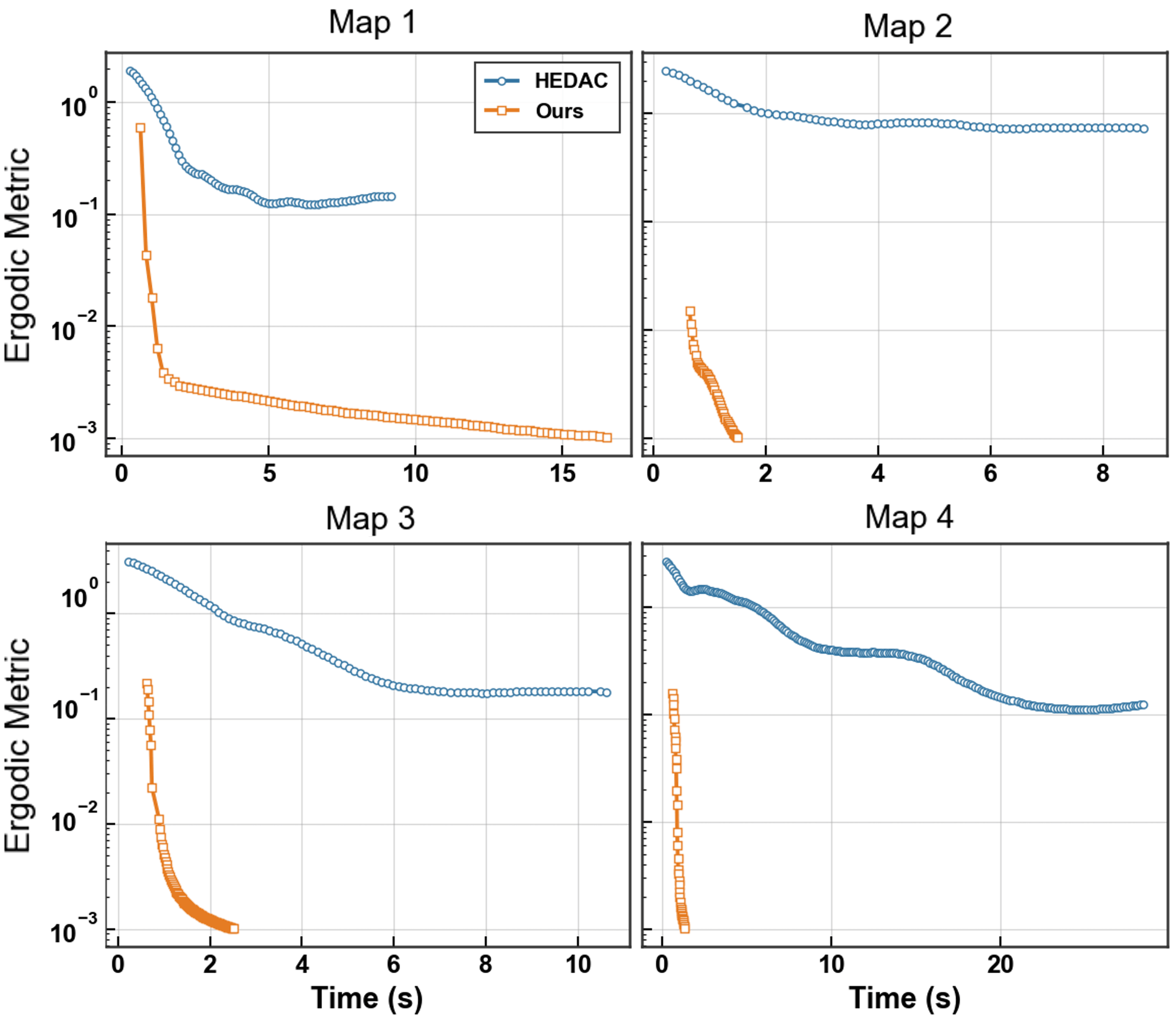}
    \caption{ The iterations of HEDAC optimized over four maps shown in Fig.~\ref{fig: Different Maps}, the footprint size is similar to the Baseline Fixed Footprint \ref{fixed_base}.}
    \label{fig:hdeac}
\end{figure}

As shown in Fig.~\ref{fig:hdeac}, baseline 4 HEDAC returns the first solution very quickly and is able to improve the solution quality as the runtime increases.
However, this improvement plateaus after a while.
In contrast, \abbrMethod achieves a two-order-of-magnitude lower ergodicity than HEDAC within 2 seconds across all four maps. Specifically, in Maps $2,3,4$, our \abbrMethod reaches the $10^{-3}$ ergodicity, which is the termination threshold for the optimization, within 2 seconds.
In Map 1, while both methods show slower convergence, \abbrMethod still reduces the ergodicity by about two orders of magnitude within 2 seconds.
Our observation aligns with the results reported in the literature \cite{dong2025ergodic}, which verifies the benefits of trajectory optimization over PDE-driven solution.

\subsection{Runtime Comparison and Ablation Study} \label{CCRA}
\label{sec:comp_analysis}

\subsubsection{Runtime Comparison against Baselines}
Although our method optimizes an additional state dimension 
(i.e., altitude in 3D), it often requires less runtime 
than all baselines across the four information maps 
(Fig.~\ref{fig: Ergo}(b)(I), \ref{fig: Ergo}(b)(II), and \ref{fig: Ergo}(b)(III)). 
Note that, the number of iterations is not always smaller. 
In certain cases (e.g., Map~1), ours requires more iterations 
than the baselines (Fig.~\ref{fig: Ergo}(c)). 
However, the total runtime remains much lower, indicating that per-iteration computational runtime is an important factor.
The reason for such computational efficiency of our \abbrMethod is due to (i) the proposed reuse of Jacobians and (ii) the proposed log-spaced line search, during the trajectory optimization.
We verify the benefits of these two proposed techniques in the following ablation study.

\subsubsection{Ablation Study}

Within our \abbrMethod framework, we consider the following variants for comparison.
\begin{itemize}

\item Armijo-NoReuse: This variant uses the popular Armijo backtracking line search approach and use the auto-differentiation provided by JAX rather than using our proposed Jacobian reuse techniques.

\item Log-NoReuse: This variant uses our log-spaced line search and use the auto differentiation provided by JAX rather than using our proposed Jacobian reuse techniques.

\item Armijo-Reuse: This variant uses the popular Armijo backtracking line search approach and use our Jacobian reuse techniques.

\item Log-Reuse: This variant is the full version of our approach, including both our log-spaced line search and our proposed Jacobian reuse techniques.

\end{itemize}

\begin{figure}[tb]
    \centering
    \includegraphics[width=\linewidth]{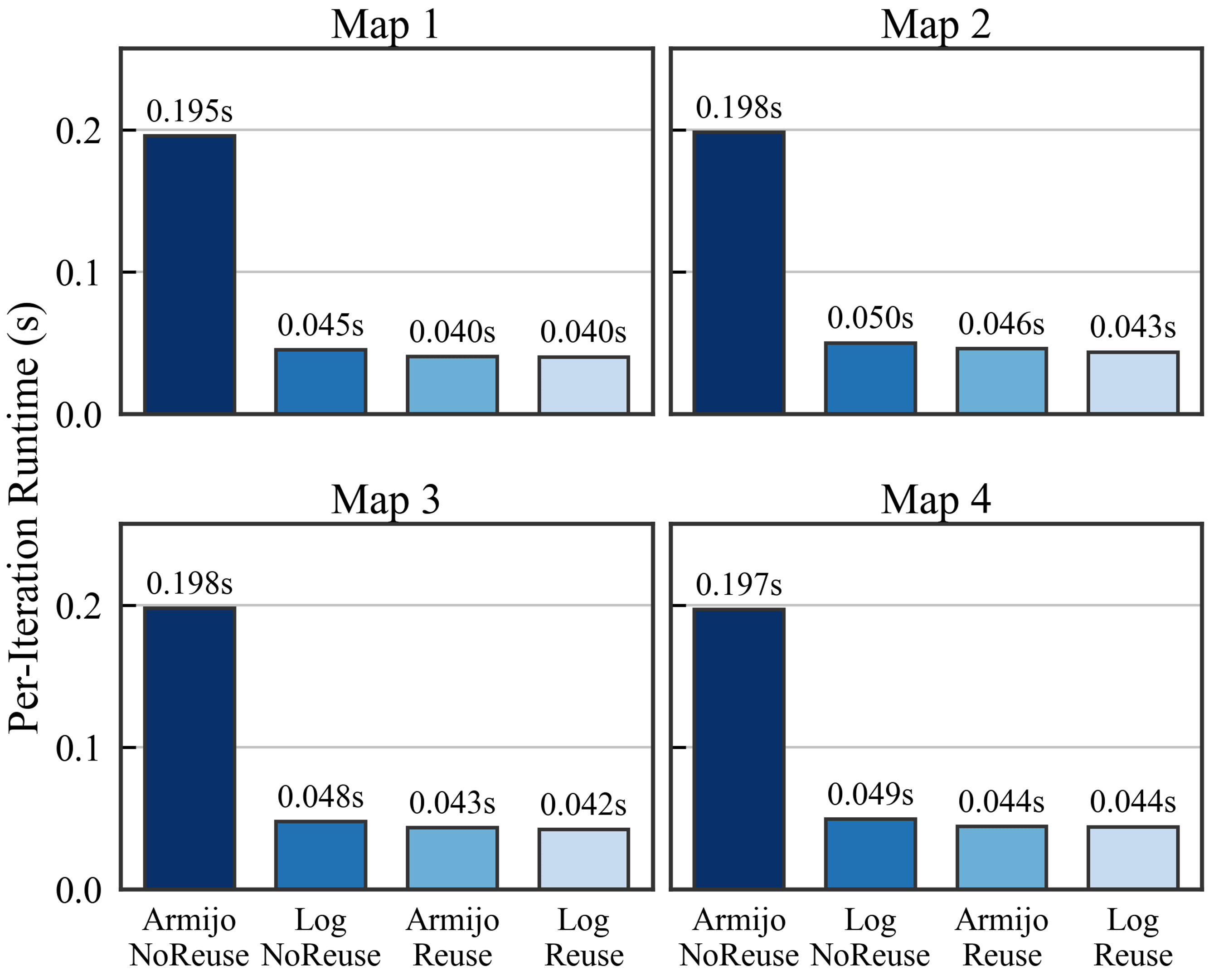}
    \vspace{-3mm}
    \caption{Per-iteration runtime comparison of different line-search strategies across four information maps shown in Fig~\ref{fig: Different Maps}. The proposed gradient reuse method consistently reduces computational cost.}
    \label{fig:2Dreuse}
\end{figure}

Fig.~\ref{fig:2Dreuse} shows the per-iteration runtime on four maps.
First, by comparing Armijo-NoReuse and Log-NoReuse, we find that the proposed log-spaced line search itself can obviously speed up the computation independently.
Second, by comparing Armijo-NoReuse and Armijo-Reuse, we find that the proposed Jacobian reuse technique itself can also obviously speed up the computation independently.
Finally, when applying both proposed techniques together, further improvement in runtime can be observed sometimes, but it is often tiny.
A possible reason is that, after the log-spaced line search reduces the number of iterations, using Jacobian reuse to further save the computation per iteration become less important since the number of iterations is already very small.

\subsection{Selection of Parameters}
\begin{figure}[tb]
    \centering
    \includegraphics[width=\linewidth,trim=2 2 2 2,clip]{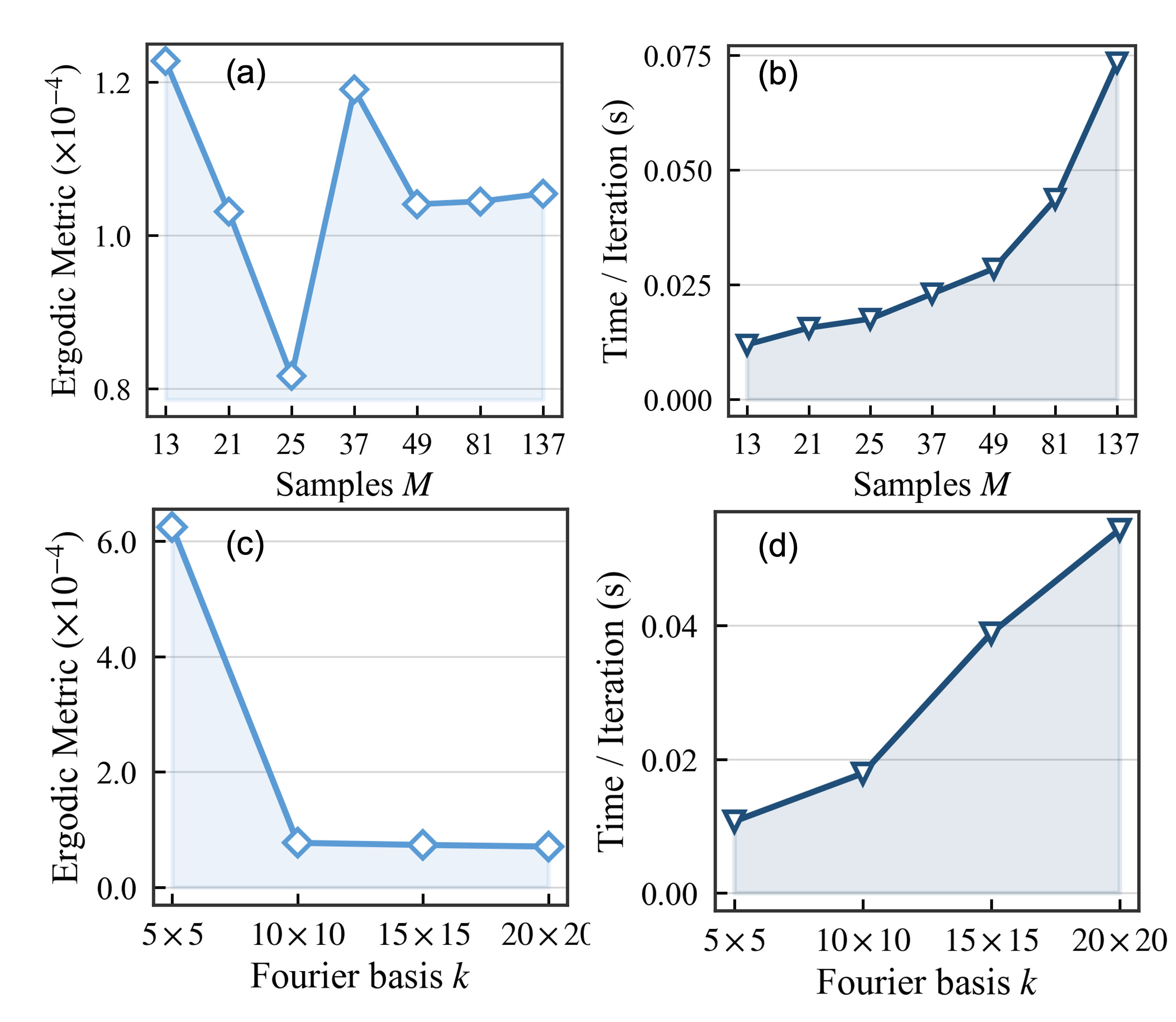}
    \vspace{-5mm}
    \caption{Ergodicity and runtime with varying $M$ (number of sampled points in the sensor footprint), and $k$ (number of Fourier basis).
    All tests are run on Map 2 in Fig.~\ref{fig: Different Maps}.
    }
    \label{fig:ES}
\end{figure}

This section studies the effect of the number of footprint samples $M$ and the number of Fourier
bases $\mathbf{k}$, using Map~2 as an example. As shown in Fig.~\ref{fig:ES}, increasing $M$ does not
consistently improve the ergodic metric while it does increase runtime, whereas increasing
$\mathbf{k}$ improves the metric with diminishing returns.

\begin{figure}[tb]
    \centering
    \includegraphics[width=\linewidth,trim=14 14 14 14,clip]{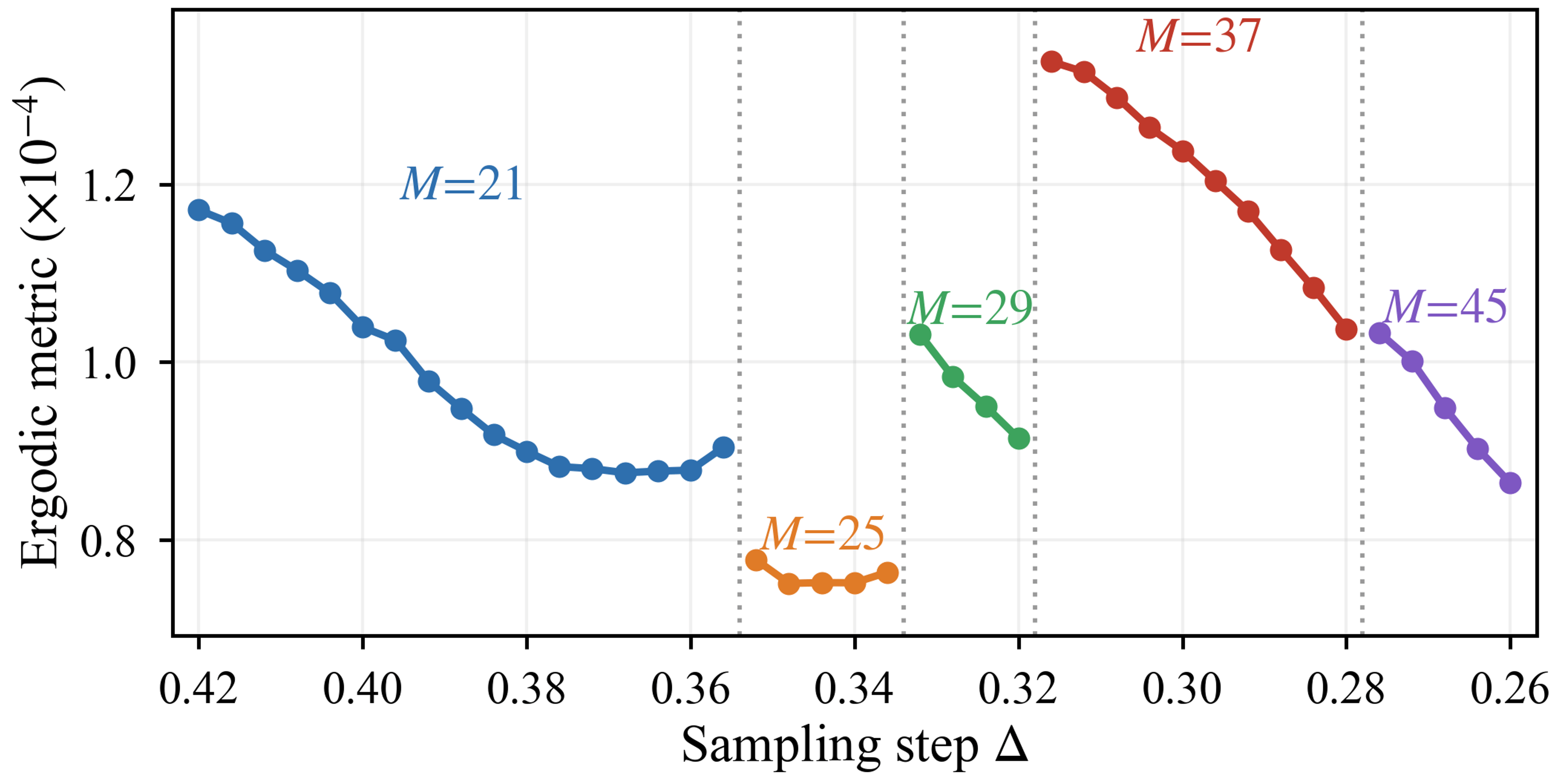}
    \vspace{-5mm}
    \caption{Footprint ergodic metric $\mathcal{E}_\gamma$ versus the footprint sampling step size $\Delta$ on Map~2.
    $\mathcal{E}_\gamma$ varies ``smoothly'' with $\Delta$ for each $M$, but jumps when $M$ changes. The reason is illustrated with Fig.~\ref{fig:seen_by_metric}.}
    \label{fig:erg_vs_delta}
\end{figure}

\begin{figure}[tb]
    \centering
    \includegraphics[width=\linewidth,trim=12 12 12 12,clip]{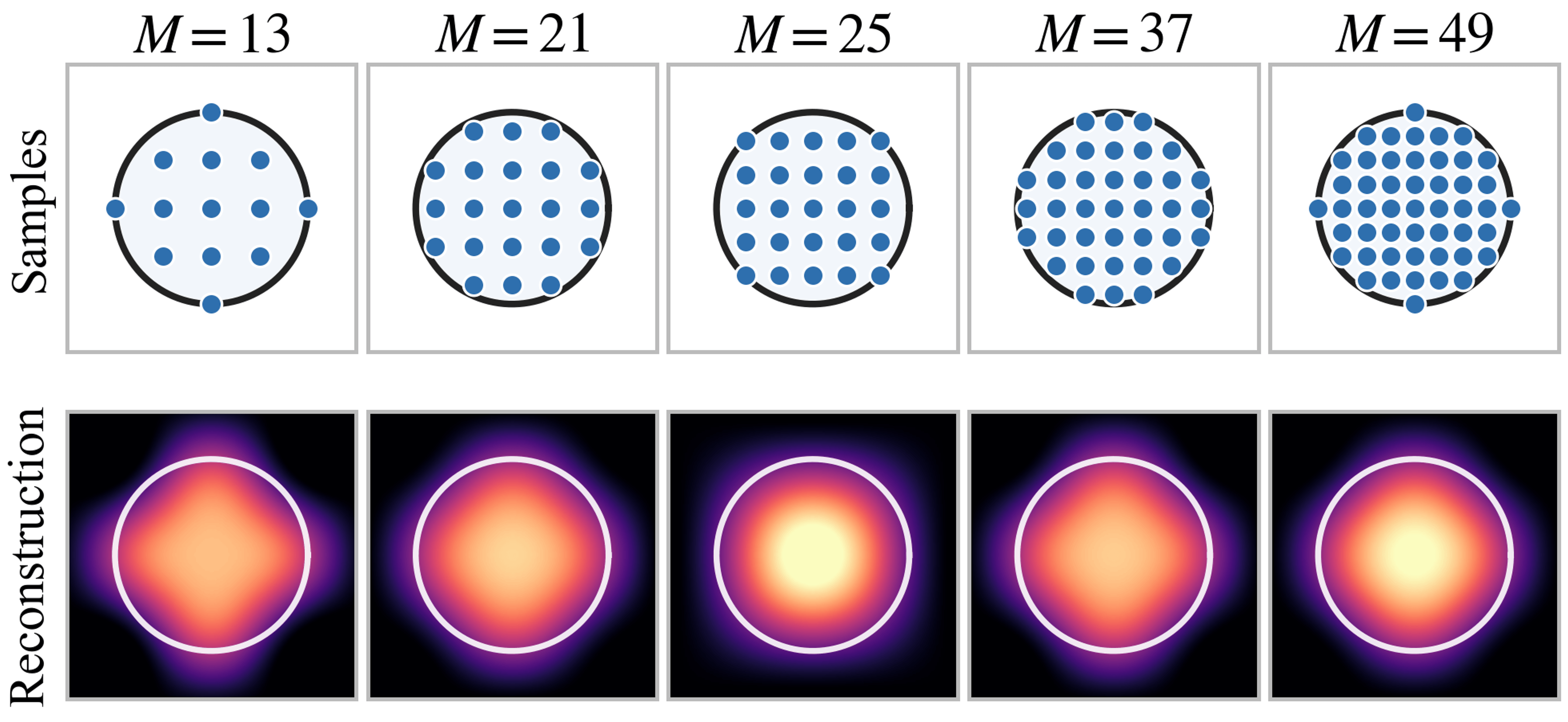}
    \vspace{-5mm}
    \caption{Reconstructed footprints for different number of sampled footprint points $M$.
    The top row shows the Cartesian grid samples retained inside the circular footprint. The bottom row shows the footprint
    reconstructed from its Fourier coefficients, where the white circle is the boundary of the footprint.
    It help illustrate the reason behind the jump in $\mathcal{E}_\gamma$ in Fig.~\ref{fig:erg_vs_delta}.
    }
    \label{fig:seen_by_metric}
\end{figure}

We find that $\mathcal{E}_{\gamma}$ has non-monotonic dependence on $M$, which is caused by the sampling of footprint points:
As shown in Fig.~\ref{fig:seen_by_metric}, the footprint is sampled on a Cartesian grid of step $\Delta$, and $M$ is the number of grid nodes falling inside the circular footprint. 
As $\Delta$ decreases, $M$ changes with discrete jumps ($13,21,25,37,\dots$), each with a different layout rather than a refinement of the previous one. Consequently, for each specific $M$, $\mathcal{E}_\gamma$ varies smoothly with $\Delta$, while $\mathcal{E}_{\gamma}$ jumps when $M$ changes (Fig.~\ref{fig:erg_vs_delta}). 
For example, the reconstructed map using Fourier coefficients for $M=25$ is circular, while varies with protruding corners at $M=37$.
Despite these jumps, note that, this variation is about $10^{-5}$, which is orders of magnitude smaller than the differences in $\mathcal{E}_\gamma$ returned by different algorithms (Fig.~\ref{fig: Ergo}).
Based on these results, we set $M=25$ and $\mathbf{k}=10\times10$ in all experiments, balancing solution quality and per-iteration runtime cost.

\subsection{Extensions to Complex Terrains}\label{Terrains_3D}
While the previous sections focused on planar settings, this section evaluates our \abbrMethod on diverse terrains to investigate:
\begin{itemize}
    \item the benefits of pose-aware planning;
    \item the computational saving of Jacobian reuse in 3D;
    \item the generalization to various 3D terrain and objects.
\end{itemize}

\subsubsection{Experiment Settings}
For experiments with 3D terrain information maps, the robot state including positions ($x$,$y$,$z$), velocities ($v_x$,$v_y$,$v_z$), and pitch and yaw ($\theta_p,\theta_i$). The robot control includes accelerations $(a_x,a_y,a_z)$ and angular velocity of pitch and yaw $(w_p,w_i)$.
The position dynamics is double integrator and the angles follow a single integrator.
The position of the robot is limited to a bounded space $[0,10]^2 \times[2,8]$, with the additional constraint that the altitude must maintain at least $0.8$ m clearance above the terrain surface.
The controls are bounded as $a_{x}, a_{y} \in [-1.5, 1.5]$ m/s$^2$, $a_z \in [-1.0, 1.0]$ m/s$^2$, and $\omega_p, \omega_y \in [-0.3, 0.3]$ rad/s.
The velocity magnitude is constrained to $\|\mathbf{v}\| \leq 3.0$ m/s, and the pitch angle is limited to $\theta_p \in [-45°, 45°]$.
We adopt the view cone defined in Sec.~\ref{def:view_cone} with half FOV angle $\alpha = 15^{\circ}$.
The footprint are computed with $N_{\text{radial}} \times N_{\text{azimuthal}} = 3 \times 12 = 36$ sampled rays plus the optical axis itself, yielding $37$ sample points.
The information maps $\phi$ are defined on the 2D terrain manifold as mentioned in Sec.~\ref{case1}. The terrain is defined as a height function $H: [0,10]^2 \to \mathbb{R}$.
As shown in the first row of Fig.~\ref{fig:traj}, the terrains are generated by mixtures of Gaussians (a-d) and a real-world terrain (e) constructed from publicly available elevation data of the Los Angeles region.
The information map are also generated with mixtures of Gaussians randomly sampled, which are then projected onto the terrain, with the red areas indicating the high-information regions as shown in the figure.
We refer to these five maps as (a) Double-Peak, (b) Ridge, (c) Valley, (d) Multi-Peak, and (e) Real-World.

\subsubsection{Pose and Position Planning}
\begin{figure}[tb]
    \centering
    \includegraphics[width=\linewidth]{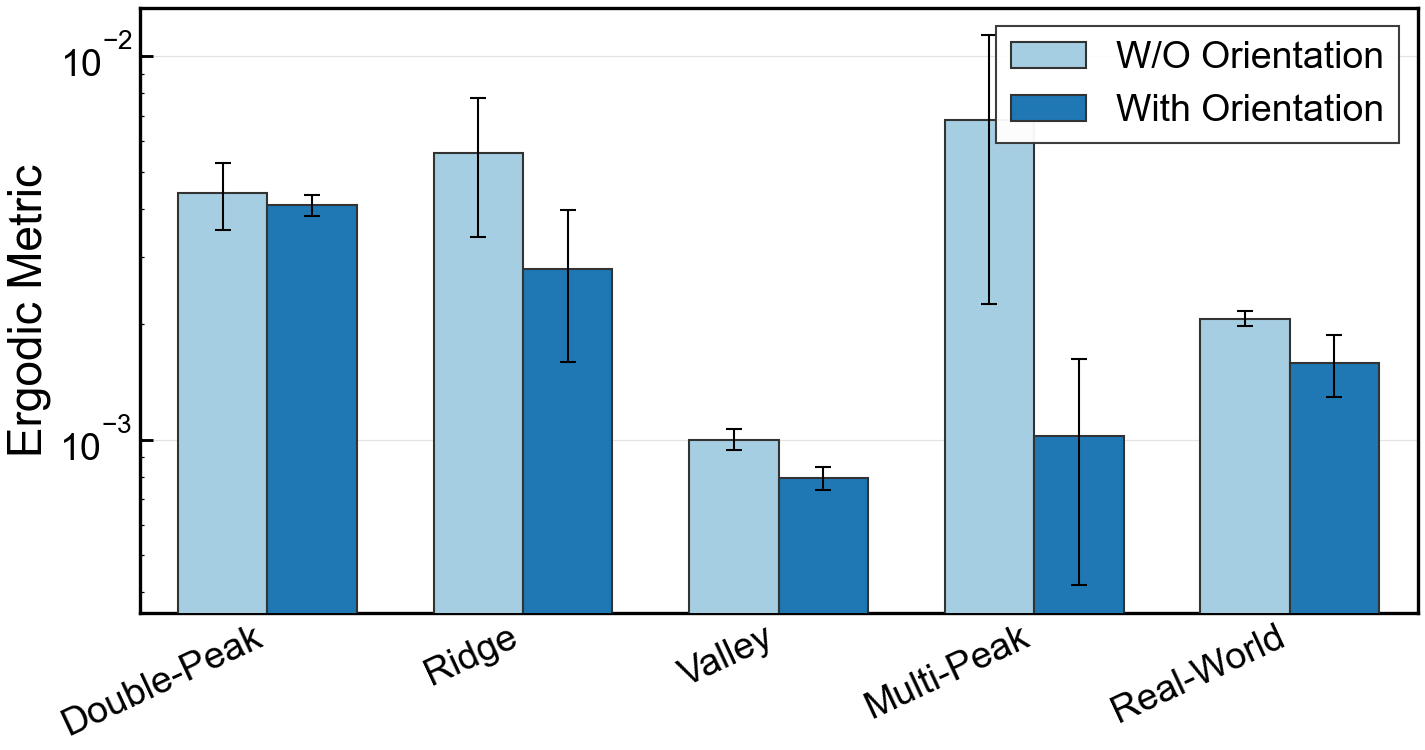}
    \vspace{-5mm}
    \caption{ The comparison between position control and altitude control in different terrains.}
    \label{fig: pose}
\end{figure}
\begin{figure*}[tb]
    \centering
    \includegraphics[width=\textwidth]{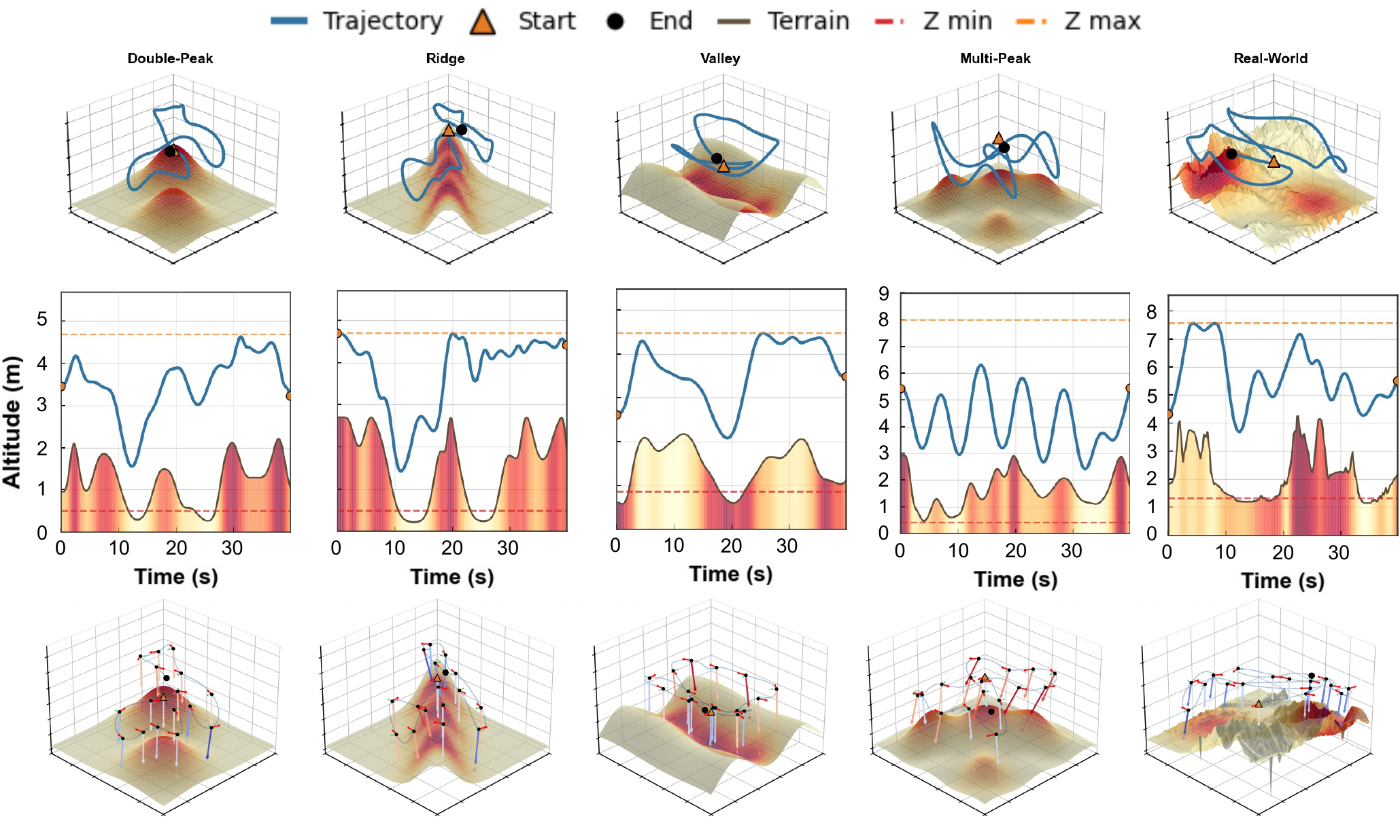}
    \vspace{-2mm}
    \caption{ SE(3) trajectories planned by our \abbrMethod over various complex terrains.
Each row corresponds to a different terrain map: (a) Double-Peak, (b) Ridge, (c) Valley,
(d) Multi-Peak, and (e) Real-World.
Among them, (e) is constructed from publicly available elevation data of the Los Angeles region.
First Row: spatial trajectories over terrain surfaces.
Middle: altitude profiles over time, including terrain height and safety bounds.
Last Row: full SE(3) pose evolution with orientation visualization.
Our \abbrMethod adjusts both position and orientation to conform to terrain geometry using its dynamic sensor footprint.
}
    \label{fig:traj}
\end{figure*}

Fig.~\ref{fig:traj} shows the optimized SE(3) trajectories over five different 3D terrains.
The first row displays the trajectories to cover the terrain.
The middle row shows the altitude over time, including terrain height and safety bounds.
The last row illustrates the poses, where arrows indicate the sensor orientation.
Intuitively, the robot tends to fly with lower altitude to move closer to the terrain in high-information region, while flying further away with a large FOV for low-information regions.
However, this is not always the case since the robot FOV has orientation and it is possible that, the robot is at a high altitude and the FOV is oblique to better cover the nearby terrains as opposed to covering the terrains beneath the robot.

To better understand the benefits of planning with orientation, we compare against a variant where the robot orientation is fixed to let the sensor always point downwards (i.e., w/o orientation).
Fig.~\ref{fig: pose} shows that, our method achieves nearly $40\%$ on average smaller (better) ergodicity by planning with orientation than w/o orientation among five different terrains.

\subsubsection{Jacobian Reuse in 3D}
To investigate the benefits of computational reuse in 3D, we compare three variants: (i) Log-NoReuse-JAX uses the automatic differentiation toolbox to compute the Jacobians; (ii) Log-NoReuse-Analytic uses the analytical Jacobian we derived without any computational reuse; and (iii) Log-Reuse-Analytic uses the analytical Jacobian we derived while leverage our proposed computational reuse as in Sec.~\ref{sec:3d_jac_reuse}. 
Note that all three variants compared here employ the same log-spaced line search to eliminate the influence of step-size selection, and the only difference lies in their gradient evaluation.

\begin{figure}[tb]
    \centering
    \includegraphics[width=\linewidth,trim=2 2 2 2,clip]{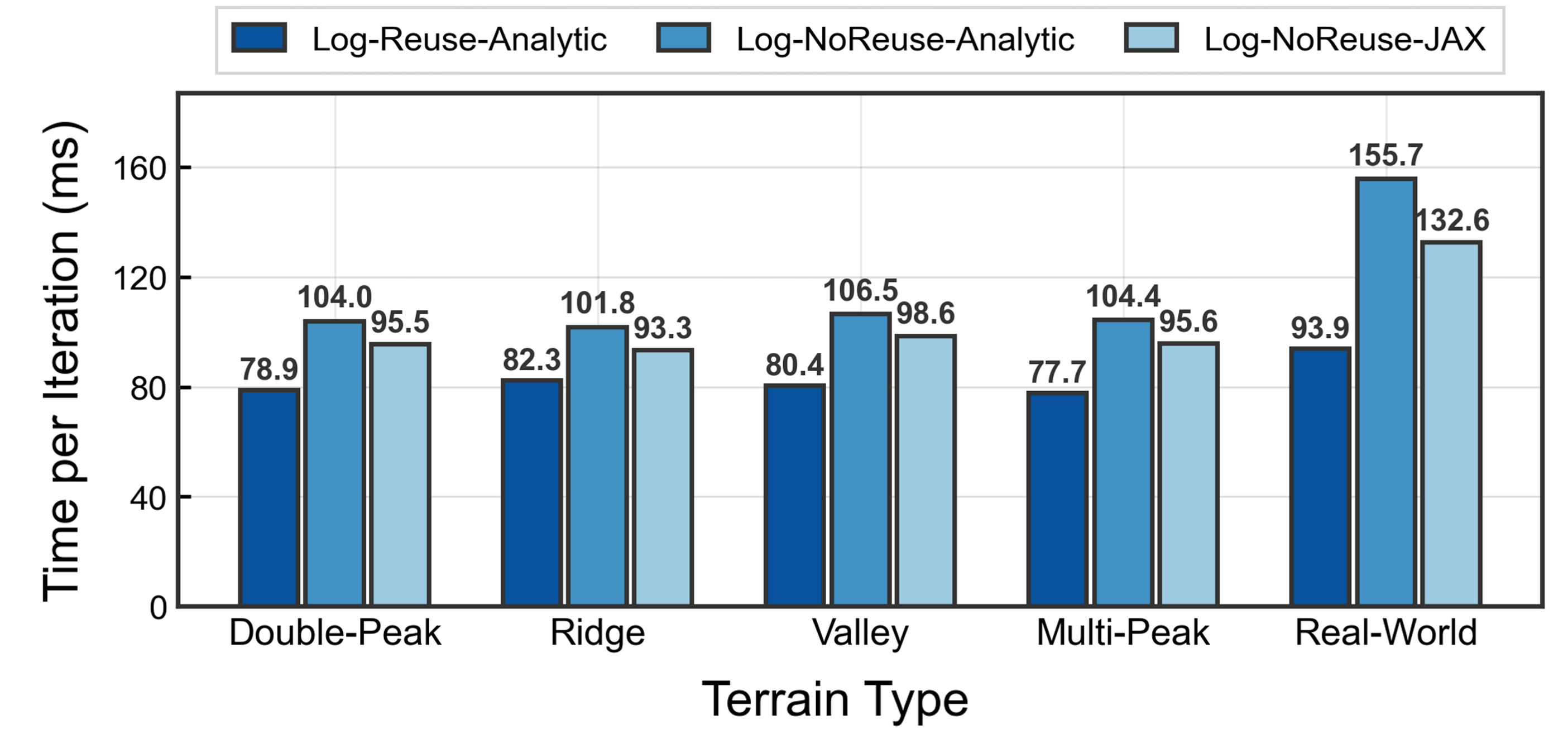}
    \vspace{-2mm}
    \caption{Per-iteration runtime for complex terrains. All planners use log-spaced line search. Our method with analytical gradient reuse can consistently reduce runtime in all terrains.
}
    \label{fig:3Dreuse_t}
\end{figure}

Fig.~\ref{fig:3Dreuse_t} shows the average per-iteration runtime in 3D terrains.
First, Log-NoReuse-Analytic is slower than Log-NoReuse-JAX, since JAX is a fast auto-diff library with implementation optimization, while Log-NoReuse-Analytic is implemented in Python and may run slow.
Second, although our Log-Reuse-Analytic is also implemented in Python directly, compared with Log-NoReuse-JAX, our Log-Reuse-Analytic reduces the per-iteration runtime by approximately $25\%$–$40\%$.
For example, in Real-World map, the per-iteration runtime is reduced from $152.3$\,ms (JAX) to $91.6$\,ms (ours), corresponding to a reduction of approximately $40$\%.

\subsection{Generalization to 3D Objects}\label{3Dreal}
While our main focus is to cover 2D manifolds, such as uneven terrains, this section evaluates the extension of our approach to cover complex objects in 3D. 
As aforementioned in Sec.~\ref{sec:tangent_error}, to avoid metric distortion, we follow the intrinsic surface representation of~\cite{dong2025ergodic}, where the ergodic metric is computed using Laplace-Beltrami eigenfunctions~\cite{Sharp:2020:LNT, 1631196} defined on the object surface mesh, rather than the flat 2D Fourier basis.
We use~\cite{dong2025ergodic} as the baseline and follow its 3D object coverage optimization setting such as the point-mass single-integrator dynamics.

For our method, we introduce two variants (ours-1, ours-2):
Ours-1 is same as the baseline with the only difference on its sensor footprint, which is no longer a fixed Gaussian.
Instead, the sensor footprint is a dynamic cone, whose optical axis points from the point-mass robot to the nearest point on the object, and the size of the cone varies based on the distance to the object. 
Ours-2 includes orientations into the robot state and dynamics, as opposed to treating the robot as a point mass as the baseline and ours-1 do.
Ours-2 aligns the optical axis with the x-axis of the robot.
We set a terminating threshold of $2\times 10^{-6}$ for ergodicity.

\begin{figure}[tb]
    \centering
    \includegraphics[width=0.85\linewidth,trim=1 1 1 1,clip]{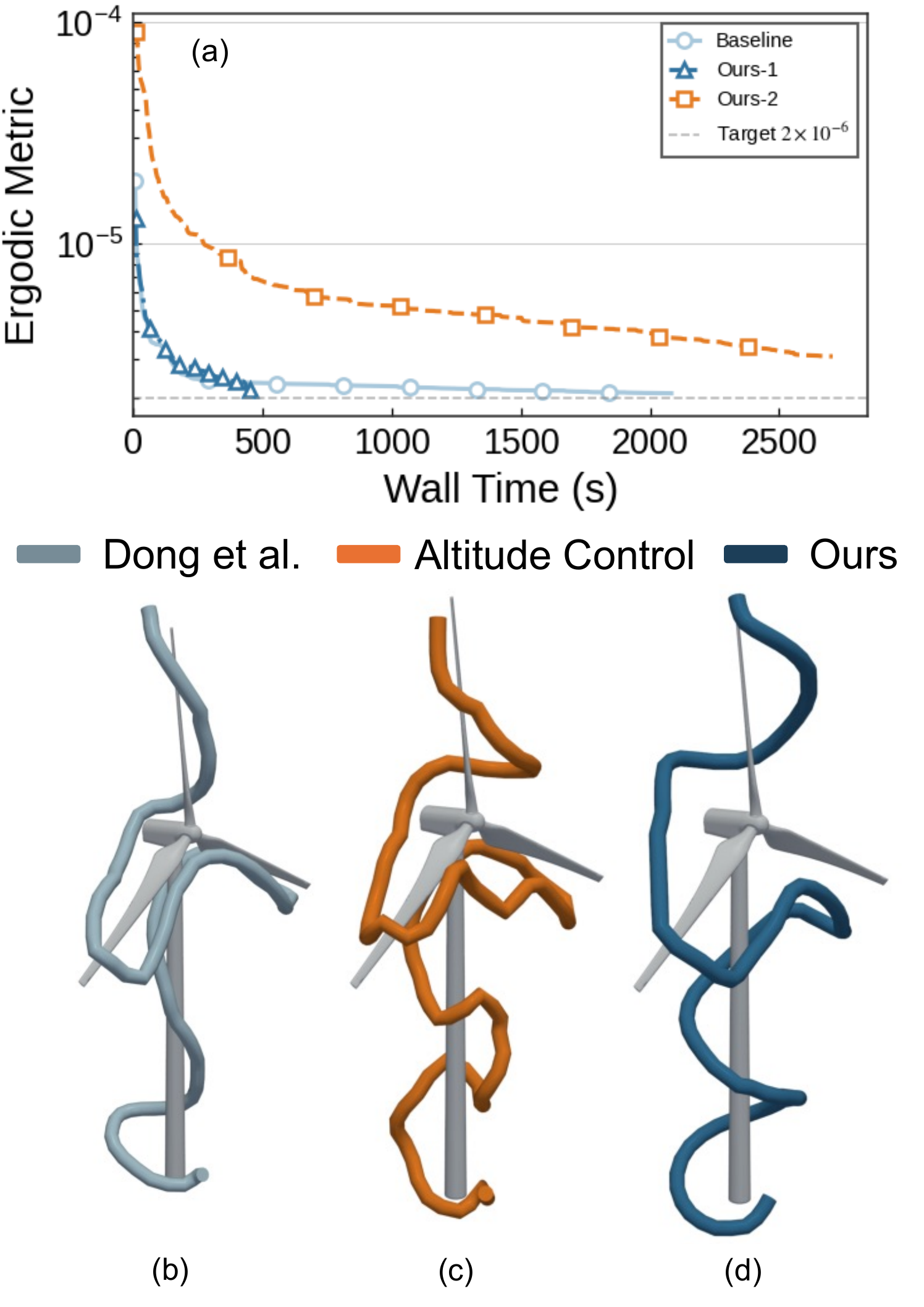}
    \caption{Tests on 3D objects. (a) Ergodic metric versus wall time. Our \abbrMethod achieves the fastest convergence and the lowest final metric. (b)-(d) Trajectory of three different methods. 
    }
    \label{fig: 3Dtraj}
\end{figure}

Fig.~\ref{fig: 3Dtraj} shows that, ours-1 and the baseline \cite{dong2025ergodic} achieve similar performance and trajectories.
Specifically, while the baseline \cite{dong2025ergodic} converges faster than ours-1 at the beginning (e.g. $< 300$ s), ours-1 eventually converges to the lowest ergodic metric very quickly among the three planners ($1.99 \times10^{-6}$ in about $500$ s), roughly $1500$ s faster than the baseline ($2.10 \times 10^{-6}$ in more than $2000$ s).
In contrast, ours-2 converges slower and yields higher ergodicity ($3.08 \times 10^{-6}$ in $2500$ s) due to the inclusion of orientation into the robot dynamics, which makes the optimization harder.
For a more rigorous discussion of ergodicity on curved surfaces, please refer to \cite{dong2025ergodic, bilaloglu2024tactile}.

\subsection{Real Robots Experiments}\label{Experiment}
We deploy our planner on quadrotors with dynamics~\cite{luis2016designtrajectorytrackingcontroller}.
We use Crazyflie 2.1 micro aerial vehicles \cite{8046794} within a motion capture system, which are controlled using the Crazyswarm2 framework \cite{preiss2017crazyswarm} integrated with ROS2 \cite{macenski2022robot}.
Each quadrotor is equipped with a downward-facing LED to visualize its sensor footprint during the robot motion.

\begin{figure}[tb]
\centering
\includegraphics[width=\linewidth,trim=2 2 2 2,clip]{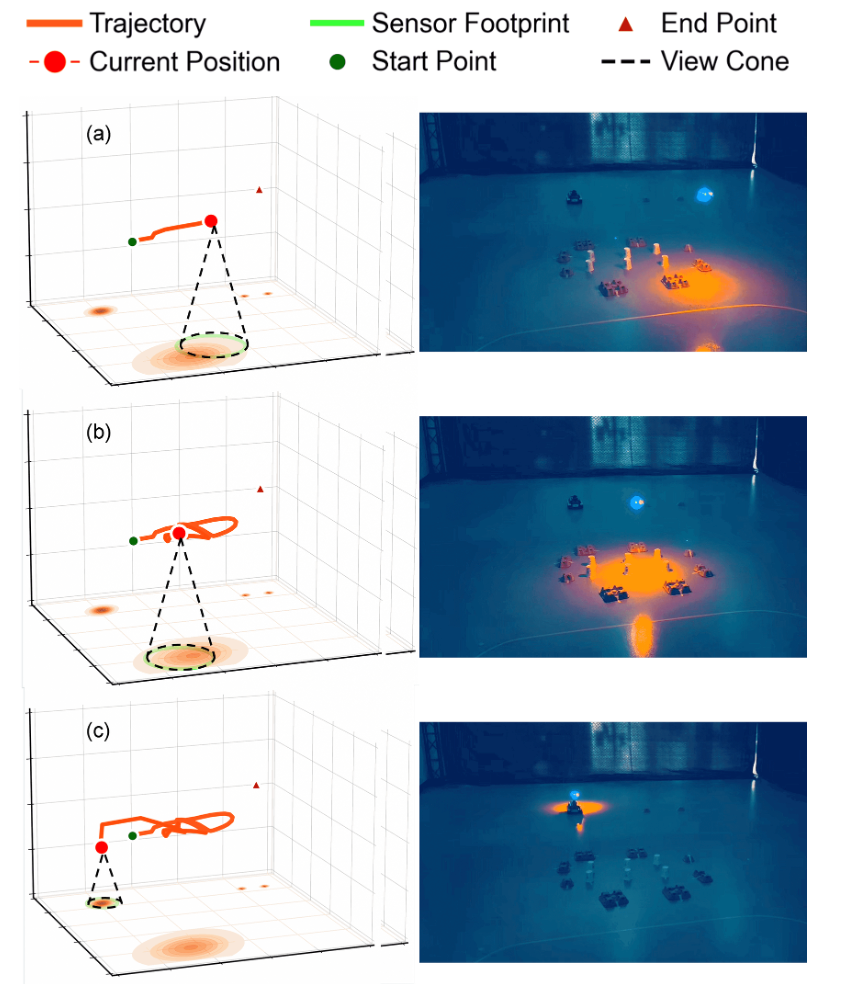}
\caption{Snapshots of a drone executing the trajectory planned by \abbrMethod in Map 1 shown in Fig. \ref{fig: Different Maps}. The drone is equipped with a downward-pointing LED to illustrate the sensor footprint. (a)-(c) show the snapshot of the system at time ${1.7,8.3,9.4}$ seconds.}
\label{fig: singlesnap}
\end{figure}

The test environment is a 2D workspace of size $4.0 \times 4.0$ meters with a single drone that seeks to ergodically cover a 2D information map.
The flying height of the drone is limited to $[0,1]$ meters.
The planning horizon is $10$ seconds with each time step being $0.1$.
Fig.~\ref{fig:abs} and ~\ref{fig: singlesnap} show the test setting and some snapshots of the robot motion.
We place a few items on the ground in the test area to indicate the information distribution.
The robot first flies high to cover the wide-spread information with a coarse sensor footprint.
After a while, the robot flies to the information peak and lowers its altitude, as shown in Fig.~\ref{fig:abs}(b), in order to cover the dense information with a fine sensor footprint.

\section{Conclusions and Future Work}
\label{sec:conclusion}

This paper explores Ergodic Trajectory Optimization with Dynamic sensor Footprints (\abbrESDF) and introduces a new footprint ergodic metric that considers the dynamic sensor footprint.
We formulate a corresponding Optimal Control Problem (OCP), and develop a numerical approach \abbrMethod to compute the proposed footprint ergodic metric and leverage Lagrangian optimization to solve the problem.
The key idea in our \abbrMethod is the derivation of analytical gradient of the sampled points in the sensor footprint with respect to the robot pose in SE(3), for both flat ground and uneven terrain represented by manifolds.
We evaluate our \abbrMethod and compare it against several recent baselines in various maps, with both simulation and real quadrotor experiments.
The results verify that, our \abbrMethod can plan trajectories for the robots by considering their varying sensor footprints so as to better cover the information map ergodically.

For future work, one can consider dynamic sensor footprints in dynamically changing information maps by integrating the Bayesian filtering framework~\cite{miller2015ergodic}, or consider connectivity maintenance and communication among the robots~\cite{2025_RSS_IMEC_YongceLiu}.

\color{black}
\bibliographystyle{IEEEtran}
\bibliography{references}

\section*{Appendix}

This appendix details the proofs of lemmas in Sec.~\ref{sec:terrain_3d}.

\subsection{Proof of Lemma~\ref{lem:position_jacobian}}
\label{app:position_jac}
For a fixed tangent plane with parameters $(\mathbf{p}_0,\mathbf{n})$ and a fixed ray with parameters $(\beta,\varphi)$, recall that the footprint point associated with ray $(\beta,\varphi)$ is
$\mathbf{q}(\beta,\varphi)=\mathbf{p}+l^*(\beta,\varphi)\,\mathbf{d}(\beta,\varphi)$,
where $l^*(\beta,\varphi)=h/D(\beta,\varphi)$,
$h=\mathbf{n}^\top(\mathbf{p}-\mathbf{p}_0)$,
and $D(\beta,\varphi)=-\mathbf{n}^\top\mathbf{d}(\beta,\varphi)$.
Then,
\begin{equation}
\frac{\partial h}{\partial \mathbf{p}}
=
\frac{\partial \mathbf{n}^\top(\mathbf{p}-\mathbf{p}_0)}{\partial \mathbf{p}}
=
\mathbf{n}^\top,
\end{equation}
$D(\beta,\varphi)$ is constant, and
\begin{equation}
\frac{\partial l^*}{\partial \mathbf{p}}
=
\frac{\partial}{\partial \mathbf{p}}\left(\frac{h}{D}\right)
=
\frac{1}{D}\frac{\partial h}{\partial \mathbf{p}}
=
\frac{1}{D}\mathbf{n}^\top.
\end{equation}
Then, Eq. \eqref{eq:position_jacobian} to be proved is obtained by:
\begin{align}
\frac{\partial \mathbf{q}}{\partial \mathbf{p}}=
\frac{\partial}{\partial \mathbf{p}}\left(\mathbf{p}+l^*\mathbf{d}\right) =\mathbf{I}_3+\mathbf{d}\frac{\partial l^*}{\partial \mathbf{p}}
=\mathbf{I}_3+\frac{1}{D}\,\mathbf{d}\,\mathbf{n}^\top.
\end{align}

\subsection{Proof of Lemma~\ref{lem:orientation_jacobian}}
\subsubsection{Cone Basis Derivatives}\label{app:cone_basis}

The cone basis vectors $\mathbf{e}_1$ and $\mathbf{e}_2$ are defined in Definition~\ref{def:cone_basis}:
\begin{equation}\label{83}
\mathbf{e}_2 = -\frac{\mathbf{a} \times \mathbf{n}}{\|\mathbf{a} \times \mathbf{n}\|}, \qquad
\mathbf{e}_1 = \mathbf{e}_2 \times \mathbf{a}.
\end{equation}

We first derive the derivative of $\mathbf{e}_2$.
Let $[\mathbf{x}]_\times \in \mathbb{R}^{3\times 3}$ denote the skew-symmetric matrix such that $[\mathbf{x}]_\times \mathbf{y} = \mathbf{x} \times \mathbf{y}$.
Let $\mathbf{c} = \mathbf{a} \times \mathbf{n}$, so $\mathbf{e}_2 = -\mathbf{c}/\|\mathbf{c}\|$ with $\|\mathbf{c}\| = \sin\theta_{\mathrm{inc}}$.
With $\frac{\partial \mathbf{c}}{\partial \mathbf{a}} = -[\mathbf{n}]_\times$, we have:
\begin{equation}
\frac{\partial \mathbf{e}_2}{\partial \mathbf{a}} = \frac{1}{\sin\theta_{\mathrm{inc}}} \left( \mathbf{I}_3 - \mathbf{e}_2\,\mathbf{e}_2^\top \right) [\mathbf{n}]_\times,
\end{equation}
where $\mathbf{I}_3 - \mathbf{e}_2\,\mathbf{e}_2^\top$ is the projection onto the plane orthogonal to $\mathbf{e}_2$.
An equivalent expansion of this expression is:
\begin{equation}
\frac{\partial \mathbf{e}_2}{\partial \mathbf{a}} = \frac{1}{\sin\theta_{\mathrm{inc}}} [\mathbf{n}]_\times + \frac{1}{\sin^2\theta_{\mathrm{inc}}} \mathbf{e}_2 \mathbf{a}^\top + \frac{\cos\theta_{\mathrm{inc}}}{\sin^2\theta_{\mathrm{inc}}} \mathbf{e}_2 \mathbf{n}^\top,
\end{equation}
which can be verified by expanding $(\mathbf{I}_3 - \mathbf{e}_2 \mathbf{e}_2^\top)[\mathbf{n}]_\times$.
With $\mathbf{e}_1 = \mathbf{e}_2 \times \mathbf{a} = -[\mathbf{a}]_\times \mathbf{e}_2$, we have:
\begin{equation}\label{85}
\frac{\partial \mathbf{e}_1}{\partial \mathbf{a}} = [\mathbf{e}_2]_\times - [\mathbf{a}]_\times \frac{\partial \mathbf{e}_2}{\partial \mathbf{a}}.
\end{equation}

\subsubsection{Derivative of Ray Direction}
Combining cone basis derivatives (Eq.~\ref{83}-Eq.~\ref{85}) and Eq~\ref{eq:ray_direction} yields
\begin{equation}\label{eq:dd_da_appendix}
\frac{\partial \mathbf{d}}{\partial \mathbf{a}} = \cos\beta\, \mathbf{I}_3 + \sin\beta \cos\varphi\, \frac{\partial \mathbf{e}_1}{\partial \mathbf{a}} + \sin\beta \sin\varphi\, \frac{\partial \mathbf{e}_2}{\partial \mathbf{a}}.
\end{equation}

\subsubsection{Derivative of Denominator}
We now derive $\frac{\partial D}{\partial a}$ with $D$ defined in Eq.~\ref{eq:D_function}.
The denominator $D$ depends on $\mathbf{a}$ through $\theta_{\mathrm{inc}}$.
Based on the geometry,
\begin{equation}
\cos\theta_{\mathrm{inc}} = -\mathbf{n}^\top \mathbf{a}, \qquad \sin\theta_{\mathrm{inc}} = \sqrt{1 - (\mathbf{n}^\top \mathbf{a})^2}
\end{equation}

Taking derivatives with respect to $\mathbf{a}$ yields
\begin{equation}
\frac{\partial \cos\theta_{\mathrm{inc}}}{\partial \mathbf{a}} = \frac{\partial (-\mathbf{n}^\top \mathbf{a})}{\partial \mathbf{a}} = -\mathbf{n}
\end{equation}

For $\sin\theta_{\mathrm{inc}}$, applying the chain rule yields
\begin{align}
\frac{\partial \sin\theta_{\mathrm{inc}}}{\partial \mathbf{a}} = \frac{\partial}{\partial \mathbf{a}} \sqrt{1 - (\mathbf{n}^\top \mathbf{a})^2} \nonumber
= \frac{\cos\theta_{\mathrm{inc}}}{\sin\theta_{\mathrm{inc}}} \mathbf{n}
\end{align}

Now we can compute $\frac{\partial D}{\partial \mathbf{a}}$:
\begin{align}
\frac{\partial D}{\partial \mathbf{a}} &= \frac{\partial \cos\theta_{\mathrm{inc}}}{\partial \mathbf{a}} \cos\beta + \frac{\partial \sin\theta_{\mathrm{inc}}}{\partial \mathbf{a}} \sin\beta \cos\varphi \nonumber \\[4pt]
&= \left( -\cos\beta + \frac{\cos\theta_{\mathrm{inc}} \sin\beta \cos\varphi}{\sin\theta_{\mathrm{inc}}} \right) \mathbf{n} \nonumber \\[4pt]
&= \frac{-\sin\theta_{\mathrm{inc}} \cos\beta + \cos\theta_{\mathrm{inc}} \sin\beta \cos\varphi}{\sin\theta_{\mathrm{inc}}} \mathbf{n}
\end{align}

Let $N_u=-\sin\theta_{\mathrm{inc}}\cos\beta+\cos\theta_{\mathrm{inc}}\sin\beta\cos\varphi$, then the above expression can be written as
\begin{equation}\label{eq:dD_da_appendix}
\frac{\partial D}{\partial \mathbf{a}} = \frac{N_u}{\sin\theta_{\mathrm{inc}}} \mathbf{n}
\end{equation}

\subsubsection{Complete Orientation Jacobian}
Note that the footprint point is $\mathbf{q} = \mathbf{p} + l^* \mathbf{d}$, where $l^* = h/D$.
Since $h = \mathbf{n}^\top(\mathbf{p} - \mathbf{p}_0)$ is independent of $\mathbf{a}$, we have
\begin{equation}
\frac{\partial l^*}{\partial \mathbf{a}} = -\frac{h}{D^2} \frac{\partial D}{\partial \mathbf{a}}
\end{equation}
The complete orientation Jacobian is then:
\begin{align}
\mathbf{J}_a(\beta, \varphi) &= \frac{\partial \mathbf{q}}{\partial \mathbf{a}} = l^* \frac{\partial \mathbf{d}}{\partial \mathbf{a}} + \mathbf{d} \frac{\partial l^*}{\partial \mathbf{a}} = \frac{h}{D}\, \frac{\partial \mathbf{d}}{\partial \mathbf{a}} - \frac{h}{D^2}\, \mathbf{d} \left( \frac{\partial D}{\partial \mathbf{a}} \right)^\top,
\end{align}
where $\frac{\partial D}{\partial \mathbf{a}}$ is given by \eqref{eq:dD_da_appendix}, and $\frac{\partial \mathbf{d}}{\partial \mathbf{a}}$ is given by \eqref{eq:dd_da_appendix}, which completes the proof.

\begin{IEEEbiography}[{\includegraphics[width=1in,height=1.25in,clip,keepaspectratio]{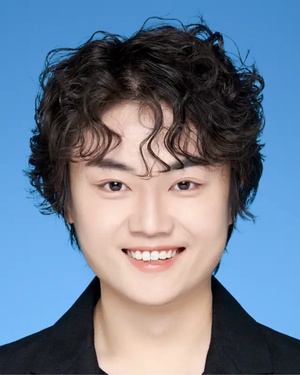}}]{Ziyue Zheng}
received the B.S. degree in optoelectronic information science and engineering from East China Normal University, Shanghai, China, in 2024. He is currently working toward the M.S. degree with the Department of Mechanical Engineering, Johns Hopkins University, Baltimore, MD, USA. His research interests include optimal control, active perception, and embodied artificial intelligence.
\end{IEEEbiography}

\begin{IEEEbiography}[{\includegraphics[width=1in,height=1.25in,clip,keepaspectratio]{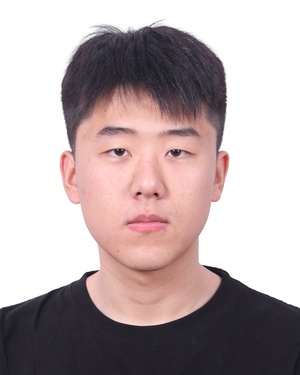}}]{Yongce Liu}
received the B.Eng. degree in automation from Northeastern University, Shenyang, China, in 2024.
He is currently working toward the M.S. degree in control science and engineering with Shanghai Jiao Tong University, Shanghai, China. His research interests include robotics, optimization, and physical artificial intelligence.
\end{IEEEbiography}

\begin{IEEEbiography}[{\includegraphics[width=1in,height=1.25in,clip,keepaspectratio]{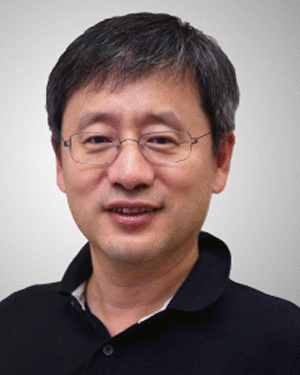}}]{Hesheng Wang}
(Senior Member, IEEE) received the B.Eng. degree in electrical engineering
from the Harbin Institute of Technology, Harbin, China, in 2002, and the
M.Phil. and Ph.D. degrees in automation and computer-aided engineering from
The Chinese University of Hong Kong, Hong Kong, in 2004 and 2007,
respectively.
He is currently a Distinguished Professor with the School of Automation and
Intelligent Sensing, Shanghai Jiao Tong University, Shanghai, China.
His current research interests include visual servoing, intelligent robotics,
computer vision, and autonomous driving.

Dr. Wang is an Associate Editor of \textit{Robotic Intelligence and
Automation} and the \textit{International Journal of Humanoid Robotics}, a
Senior Editor of the \textit{IEEE/ASME Transactions on Mechatronics}, and the
Editor-in-Chief of \textit{Robot Learning}.
He served as an Associate Editor of the \textit{IEEE Transactions on
Robotics} from 2015 to 2019 and the \textit{IEEE Transactions on Automation
Science and Engineering} from 2021 to 2023.
He was the General Chair of IEEE/RSJ IROS 2025, IEEE ROBIO 2022, and IEEE
RCAR 2016.
\end{IEEEbiography}

\begin{IEEEbiography}[{\includegraphics[width=1in,height=1.25in,clip,keepaspectratio]{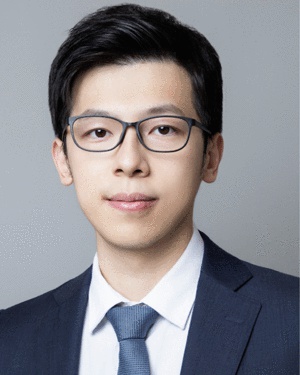}}]{Zhongqiang (Richard) Ren}
(Member, IEEE) received the dual B.E. degree in mechatronics from Tongji University, Shanghai, China, and FH Aachen University of Applied Sciences, Aachen, Germany, in 2015, and the M.S. degree in mechanical engineering and the Ph.D. degree in mechanical engineering from Carnegie Mellon University, Pittsburgh, PA, USA, in 2017 and 2022, respectively.
He is currently an Associate Professor with the Global College, Shanghai Jiao Tong University, Shanghai, China.

His research interests include robotics, path and motion planning, and multi-robot systems.
He is an Associate Editor for IEEE Transactions on Robotics, Autonomous Robots, ICRA and IROS.
\end{IEEEbiography}

\end{document}